\documentclass{article}

\PassOptionsToPackage{numbers,compress}{natbib}

\usepackage[preprint]{neurips_2026}

\usepackage[utf8]{inputenc} 
\usepackage[T1]{fontenc}    
\usepackage{hyperref}       
\usepackage{url}            
\usepackage{booktabs}       
\usepackage{amsmath}        
\usepackage{amsfonts}       
\usepackage{nicefrac}       
\usepackage{microtype}      
\usepackage{xcolor}         
\usepackage{graphicx}       
\usepackage{subcaption}     
\usepackage{wrapfig}        
\usepackage{float}          
\usepackage{makecell}       
\usepackage{multirow}       
\usepackage{listings}       

\title{LoopUS: Recasting Pretrained LLMs into Looped Latent Refinement Models}

\author{%
  Taekyhun Park \\
  Department of Data Science \\
  Pusan National University \\
  Busan, Republic of Korea \\
  \texttt{pthpark1@pusan.ac.kr}\\
  \And
  Yongjae Lee \\
  Department of  Industrial Engineering \\
  Pusan National University \\
  Busan, Republic of Korea \\
  \texttt{yongzzai1102@gmail.com}\\
  \And
  Dohee Kim \\
  Department of Artificial\\Intelligence Engineering \\
  Changwon National University \\
  Changwon, Republic of Korea \\
  \texttt{kimdohee@changwon.ac.kr}\\
  \And
  Hyerim Bae\thanks{Corresponding author} \\
  Department of  Industrial Engineering \\
  Pusan National University \\
  Busan, Republic of Korea \\
  \texttt{hrbae@pusan.ac.kr} \\
}

\usepackage{kotex}

\begin{document}

\maketitle

\begin{abstract}
  Looped computation shows promise in improving the reasoning-oriented performance of LLMs by scaling test-time compute. However, existing approaches typically require either training recurrent models from scratch or applying disruptive retrofits, which involve substantial computational costs and may compromise pretrained capabilities. To address these limitations, we introduce \textbf{Looped Depth Up-Scaling} (LoopUS), a post-training framework that converts a standard pretrained LLM into a looped architecture. As a key technical contribution, LoopUS recasts the pretrained LLM into an encoder, a looped reasoning block, and a decoder. It operationalizes this latent-refinement architecture through four core components: (1) block decomposition, guided by staged representation dynamics; (2) an input-dependent selective gate to mitigate hidden-state drift; (3) random deep supervision for memory-efficient learning over long recursive horizons; and (4) a confidence head for adaptive early exiting. Collectively, these mechanisms transform a standard non-looped model into a looped form while stabilizing it against both computational bottlenecks and representation collapse. Through stable latent looping, LoopUS improves reasoning-oriented performance without extending the generated traces or requiring recurrent training from scratch. For more details, see \url{https://thrillcrazyer.github.io/LoopUS}.
\end{abstract}

\section{Introduction}
\label{sec:intro}

The reasoning performance of large language models (LLMs) can be improved during inference by allocating additional computation, or test-time compute (TTC), in latent space to refine hidden states before producing the next token~\cite{lahoti2026mamba,wu2025inference,snell2025scaling}. By deepening internal processing rather than inflating sequence length, latent-space computation offers a complementary axis along which reasoning capacity can scale within a fixed model without increasing its parameter count~\cite{zhu2025scaling,bae2025mixtureofrecursions,geiping2025scalingtesttimecompute,gladstone2026energybased,wang2025hierarchicalreasoningmodel,nie2025large}. Looped language models are one example of this paradigm: they iterate a designated block (e.g., a transformer block or a stack of layers) to increase effective computational depth without additional parameters. However, training looped architectures from scratch is expensive at modern scales~\cite{zhu2025scaling,geiping2025scalingtesttimecompute}.

As an alternative, recent studies~\cite{mcleish2025retrofitted,bae2025relaxedrecursive} have explored tuning pretrained LLMs into a looped form. However, these approaches suffer from three limitations: 
\textit{(i)} There is no principled recipe for identifying which layers should be reused as the recurrent block because existing methods rely on heuristics rather than an analysis of the internal representation dynamics of the model~\cite{mcleish2025retrofitted,bae2025relaxedrecursive}.
\textit{(ii)} Naive iteration causes hidden-state drift because the layers were trained for single-pass use at a fixed depth rather than as a recurrent operator. Repeated reuse can therefore degrade representational fidelity, preventing iterative refinement of output quality~\cite{chen2026loopbridgeloopedtransformers}.
\textit{(iii)} Backpropagation through a long unrolled loop is both memory-intensive and prone to vanishing or exploding gradients~\cite{phd/ca/Sutskever13,pascanu2013difficulty,zhu2025surveylatentreasoning}.

\begin{figure}[t!]
  \centering
  \begin{subfigure}[b]{0.48\textwidth}
    \centering
    \includegraphics[width=\textwidth]{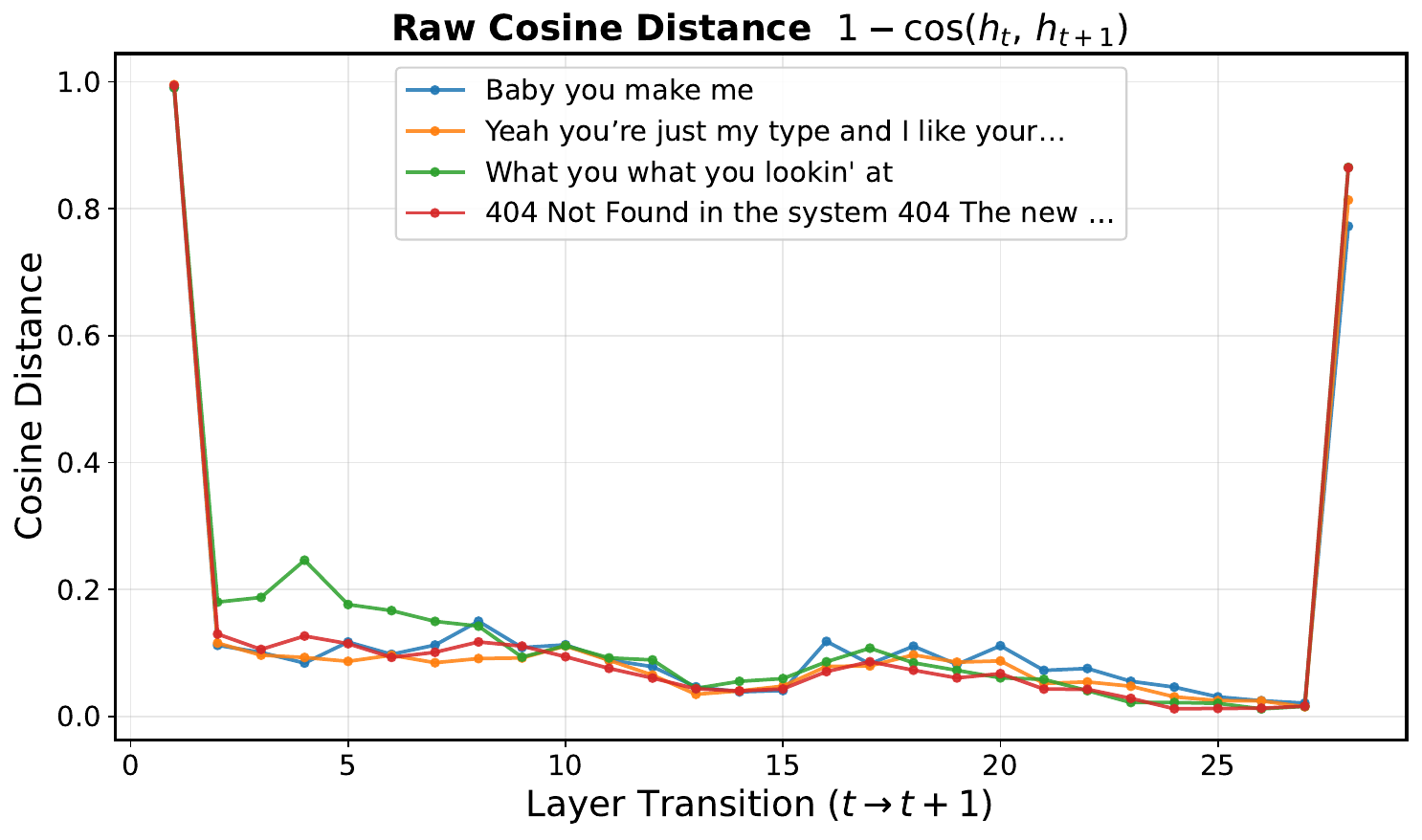}
    \caption{Hidden state distance across layers.}
    \label{fig:hidden_distance_original_LLM}
  \end{subfigure}
  \hfill
  \begin{subfigure}[b]{0.48\textwidth}
    \centering
    \includegraphics[width=\textwidth]{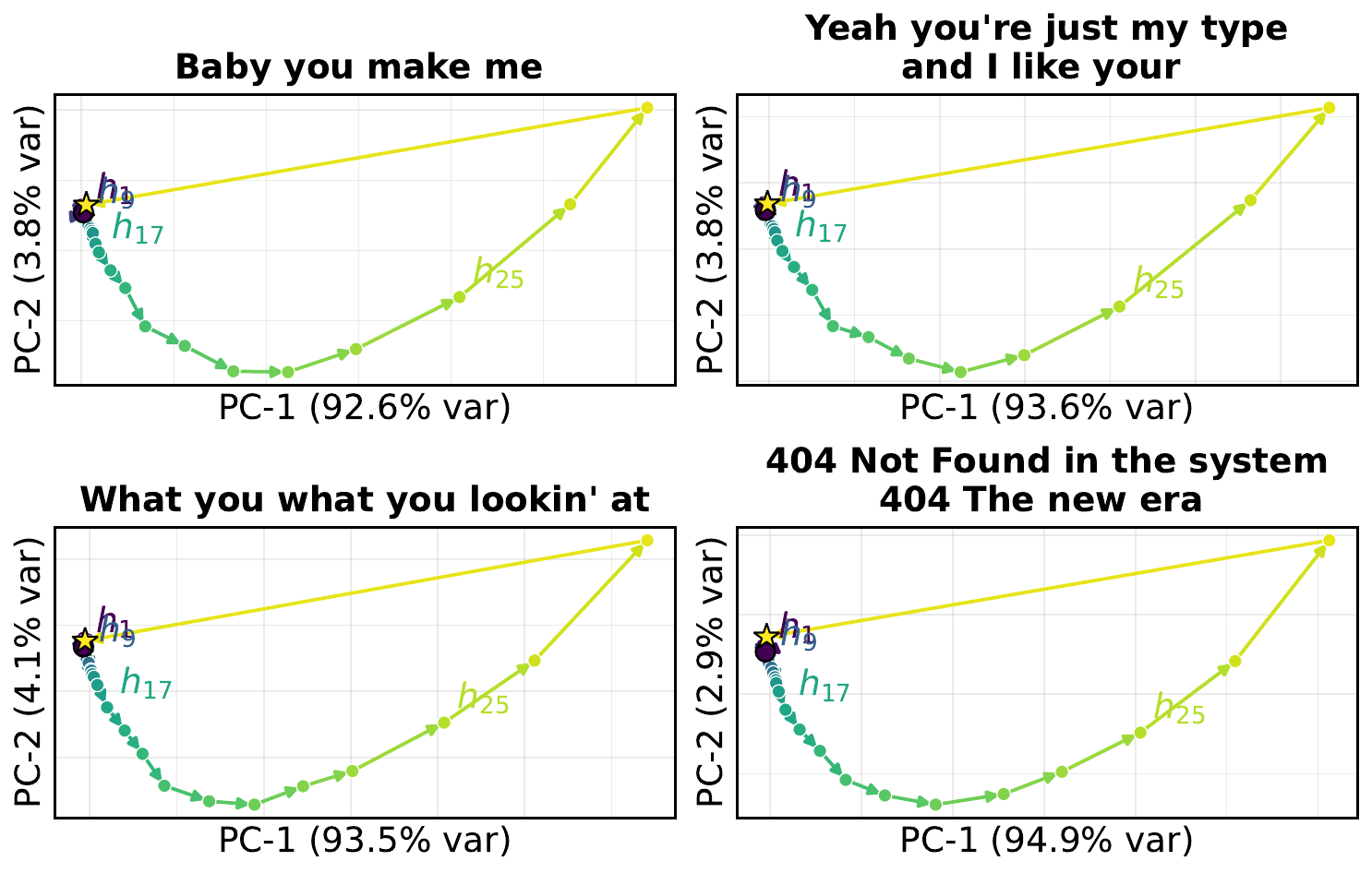}
    \caption{Hidden state trajectory visualized via PCA.}
    \label{fig:hidden_pca_original_LLM}
  \end{subfigure}
  \caption{
    Staged representation dynamics in \texttt{Qwen/Qwen3-1.7B}.
    (a) Cosine distance between consecutive hidden states reveals three distinct regimes.
    (b) Hidden-state trajectories confirm that middle layers trace a gradual arc within a confined
    region of latent space, while the final layers project sharply toward the output vocabulary space.}
  \label{fig:hidden_analysis_original_LLM}
\end{figure}

We begin by analyzing the hidden-state geometry of a pretrained LLM to understand how representations evolve across depth. In our preliminary investigation (Figure~\ref{fig:hidden_analysis_original_LLM}), the representation trajectory follows a \emph{staged pattern}: early layers rapidly transform token embeddings, middle layers evolve gradually within a stable plateau, and final layers make a sharp transition toward output decoding. This pattern is consistent with recent findings on hidden-state geometry~\cite{shibata2026suppressingfinallayerhidden, FrozenintheMiddle, ng2026rys}. Building on this observation, we decompose the LLM into three functionally distinct blocks.
 
We propose \textbf{Loop}ed Depth \textbf{U}p-\textbf{S}caling (\textbf{LoopUS}), a post-training framework that recasts a pretrained LLM into a looped form through four components. \textit{(i)} \emph{Block Decomposition} resolves the layer-selection problem by partitioning the model into encoder, reasoning, and decoder blocks, grounded in the staged representation dynamics shown in Figure~\ref{fig:hidden_analysis_original_LLM} rather than relying on heuristic layer selection. Note that only the reasoning block is reused as the loop body. \textit{(ii)} A \emph{Selective Gate} addresses hidden-state drift by interpolating each proposed update with the previous state, turning every iteration into a damped refinement step instead of an unconstrained jump. \textit{(iii)} \emph{Random Deep Supervision} sidesteps full Backpropagation Through Time (BPTT): at each step, only a few uniformly sampled iterations receive gradients, while the rest run detached. This keeps training manageable as the loop budget grows. \textit{(iv)} A \emph{Confidence Head} predicts when further refinement is unnecessary, enabling adaptive test-time compute that allocates more iterations to harder inputs and fewer to easier ones.

Empirically, LoopUS improves zero-shot accuracy by 3.0\% over pretrained backbones and reduces WikiText and LAMBADA perplexities by 17.4\% and 21.3\%, respectively. It also demonstrates high adaptation efficiency, yielding a 14.6\% relative gain on TinyLlama with 17--20$\times$ fewer training tokens than existing looped baselines. Our analyses confirm that training remains stable across extended loop depths: hidden-state trajectories contract and token distributions sharpen, indicating that gains stem from controlled, iterative latent refinement rather than uncontrolled depth expansion.

The main contributions of this paper are threefold:

\begin{itemize}
  \item \textbf{Representation-guided looped post-training framework:} We propose LoopUS, a post-training framework that converts a pretrained LLM into a looped latent-reasoning model. LoopUS decomposes the model into encoder, reasoning, and decoder blocks using staged representation dynamics, and reuses only the middle reasoning block as the loop body.
  \item \textbf{Stable and efficient latent recursion:} We introduce mechanisms that make latent looping stable and practical in pretrained LLMs, including a Mamba-inspired selective decay gate, random deep supervision, and a confidence head. The gate mitigates hidden-state drift, while random deep supervision avoids full BPTT over long recursive horizons.
  \item \textbf{Empirical analysis of loop dynamics:} We show that LoopUS improves reasoning-oriented performance, remains competitive under limited training budgets, and exhibits convergent loop dynamics through loop-depth analyses, latent-trajectory visualizations, token-level prediction analyses, and component ablations.
\end{itemize}
 
\section{Background}

\paragraph{LLM Hidden State Representations.}
Recent LLM interpretability studies, including Anthropic's work~\cite{anthropic2024mappingmind,anthropic2025biology}, suggest that LLM hidden states quickly move into an abstract predictive space in which high-level concepts can be represented, manipulated, and refined across depth rather than being rewritten at each layer. Prior studies on representation evolution and logit-lens analyses demonstrate a progression from local, lexical processing in lower layers to increasingly abstract, prediction-oriented representations in deeper layers~\cite{nostalgebraist2020logitlens,voita2019bottomup}. In this context, middle layers often form a plateau that changes relatively little, encoding information needed for the final prediction~\cite{FrozenintheMiddle}. This is followed by a sharper transition near the final layers, where representations are further transformed toward the vocabulary space~\cite{shibata2026suppressingfinallayerhidden}. Furthermore, \citet{ng2026rys} and Upstage~\cite{kim-etal-2024-solar} show that duplicating or stacking pretrained blocks can improve performance. We therefore treat the middle layers as a reusable latent workspace, exploiting this region through looping rather than by adding distinct blocks.

\paragraph{Looped LLMs.} 
Complementing TTC~\cite{wei2022chain}, which scales sequence length to elicit more explicit reasoning, looped transformers scale \emph{computation depth} by repeatedly applying the same block to refine latent representations without increasing the parameter count~\cite{mcleish2025retrofitted,zeng2025ponderlm,fu2025thinkathard,zhu2025surveylatentreasoning}. Building on recurrent-transformer formulations~\cite{dehghani2024universal}, retrofitted recurrence~\cite{bae2025relaxedrecursive,mcleish2025retrofitted,bae2025mixtureofrecursions}, latent refinement~\cite{zeng2025ponderlm,fu2025thinkathard,geiping2025scalingtesttimecompute}, and adaptive recursion~\cite{zhu2025ouro}, this work treats inference-time compute as repeated hidden-state computation. LoopUS is most similar to retrofitting-based approaches, but differs in its use of block decomposition to ground the loop, selective gating and random deep supervision to explicitly stabilize latent refinement, and a learned confidence head to enable adaptive computation.

\paragraph{Deep Learning Gating Mechanisms.}
Gating mechanisms have long been used to regulate state updates in recurrent and deep networks~\cite{hochreiter1997long,cho2014learning,srivastava2015training}. For our setting, the key distinction is between \emph{softmax-style} gating, which normalizes scores across alternatives~\cite{10264112}, and \emph{decay-style} gating, which directly controls state retention~\cite{gu2024mambalineartimesequencemodeling}. Recent sequence models increasingly adopt the latter approach, ranging from simple exponential decay to Mamba-style input-dependent selective decay~\cite{gu2022efficiently,gu2024mambalineartimesequencemodeling,beck2024xlstm,behrouz2024titans,yang2025gated}. LoopUS follows this Mamba-style perspective in the depth domain, using an input-dependent exponential decay gate that is well-suited for iterative refinement.
\section{Looped Depth Up-Scaling (LoopUS)}
\label{sec:method}
\begin{figure}[t]
  \centering
  \includegraphics[width=\textwidth]{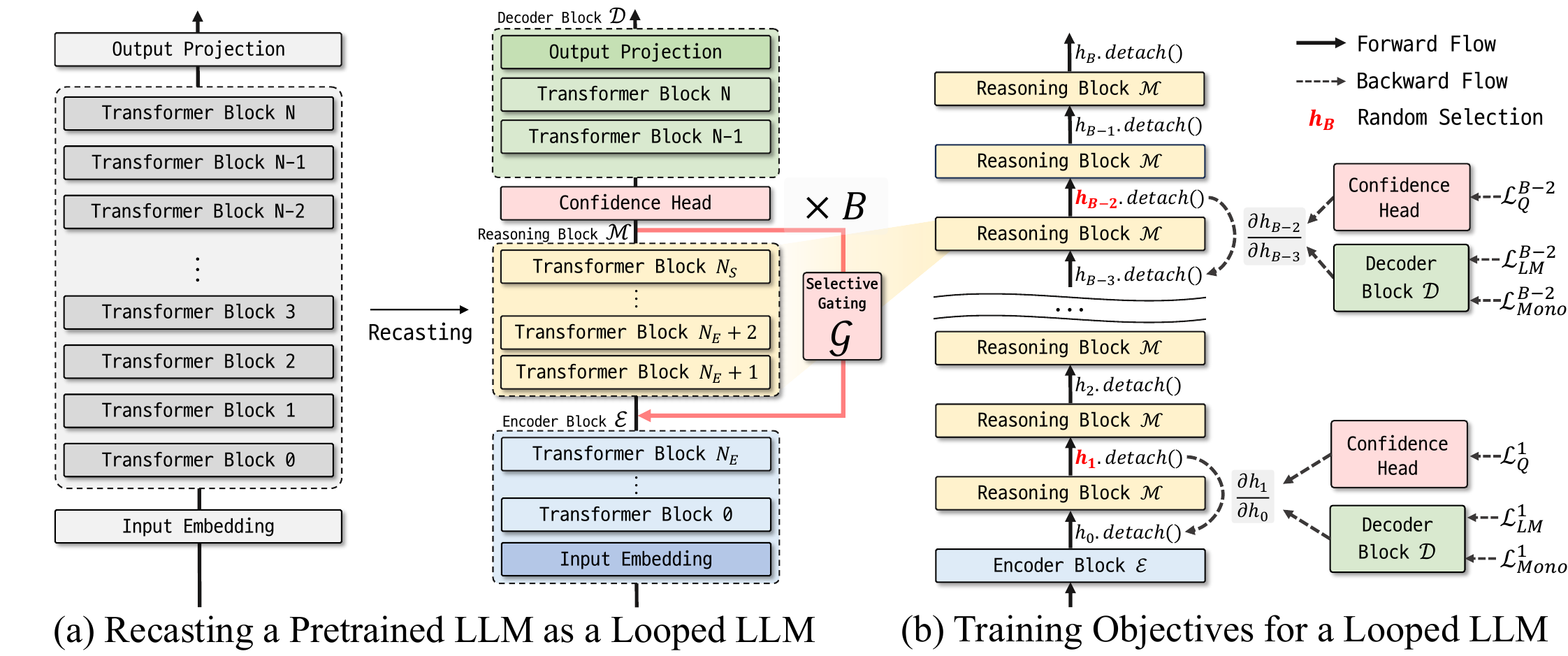}
  \caption{Overview of the LoopUS architecture.
  (a) A pretrained LLM is recast into encoder, reasoning, and decoder blocks, using a selective gate ($\mathcal{G}$) inserted between loop iterations to stabilize the loop dynamics. (b) The looped LLM is trained with random deep supervision using next-token prediction loss ($\mathcal{L}_{\mathrm{LM}}$), monotonicity loss ($\mathcal{L}_{\text{Mono}}$), and confidence loss ($\mathcal{L}_{\text{Q}}$).}
  \label{fig:framework}
\end{figure}

\subsection{Recasting LLM as a Looped LLM}
\label{sec:method_recasting}
As shown in Figure~\ref{fig:framework} (a), LoopUS partitions a pretrained LLM into an encoder $\mathcal{E}$, a reasoning block $\mathcal{M}$, and a decoder $\mathcal{D}$. Following Mi:DM~\cite{shin2026mi}, we choose this front-middle-back split based on cosine-similarity analysis across depth, placing the encoder-reasoning and reasoning-decoder boundaries near the layers where the similarity profile changes most abruptly.

Given an input sequence, LoopUS applies the encoder once to obtain the initial representation:
\begin{equation}
h^{(0)} = \mathcal{E}(x_{0:T}), \qquad h^{(0)} \in \mathbb{R}^{T \times h}.
\end{equation}
It then performs $B$ loop iterations. For $b = 0, \ldots, B-1$, the reasoning block proposes an update, and the selective gate incorporates it into the current hidden state:
\begin{equation}
\begin{aligned}
  \mathcal{R}(h^{(b)}) &= \mathcal{G}\!\left(\mathcal{M}, h^{(b)}\right), 
  h^{(b+1)} &= \mathcal{R}(h^{(b)}), 
  h^{(B)} &= \underbrace{\mathcal{R} \circ \mathcal{R} \circ \cdots \circ \mathcal{R}}_{B\ \text{iterations}} \left(h^{(0)}\right)
\end{aligned}
\end{equation}
Here $\mathcal{G}$ is introduced in Section~\ref{sec:method_gate}. After $B$ iterations, the decoder maps the final refined state to vocabulary logits, $\ell^{(B)} = \mathcal{D}(h^{(B)})$.

\paragraph{Selective Gating for Stable Loop Dynamics.}
\label{sec:method_gate}

\begin{wrapfigure}{R}{0.45\textwidth}
	\centering
	\includegraphics[width=\linewidth]{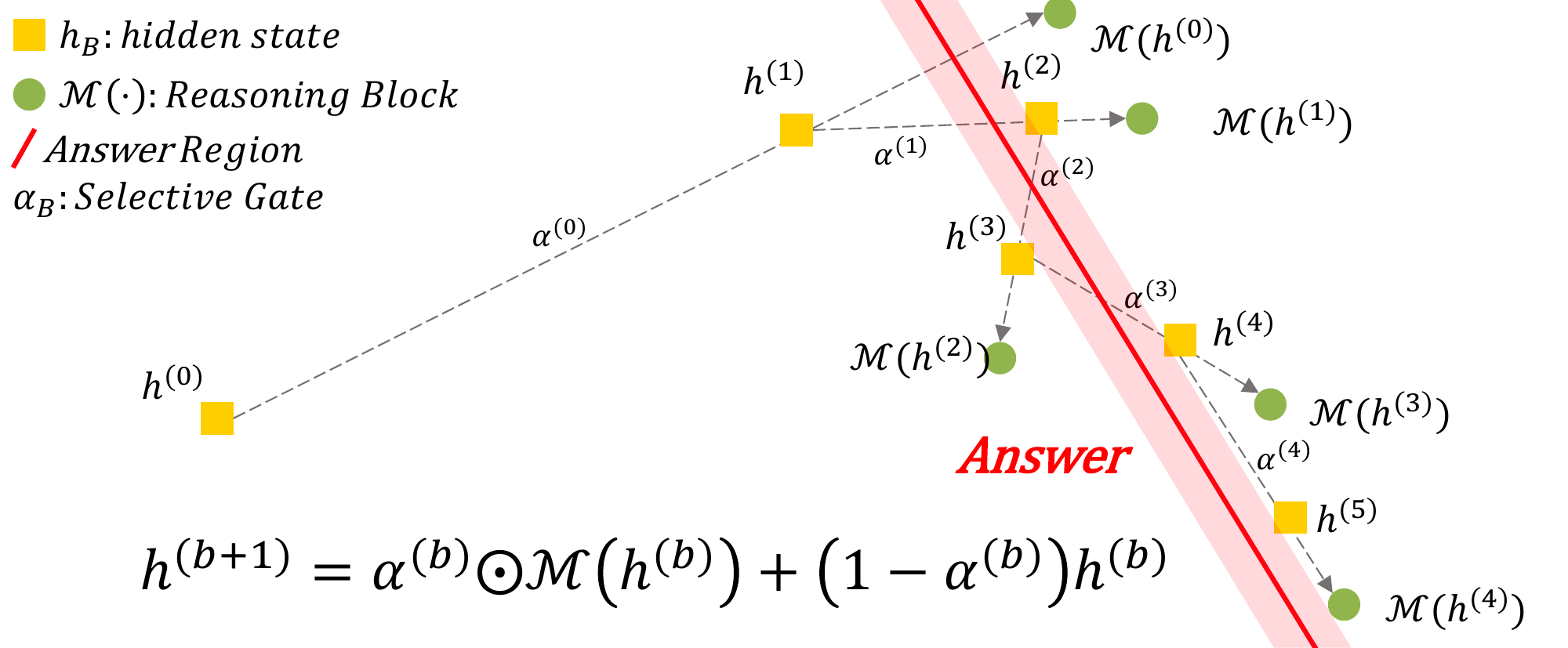}
  \caption{Conceptual view of latent refinement in LoopUS. As the reasoning block is looped, each proposed update is mixed with the previous hidden state by the selective gate, gradually steering the trajectory toward the answer region instead of allowing it to drift.}
	\label{fig:representation_learning}
\end{wrapfigure}

Naively reapplying a pretrained middle block induces hidden-state drift, as it was originally optimized for single-pass execution rather than as a recurrent operator~\cite{mcleish2025retrofitted}. Therefore, a stable latent workspace requires a structural condition restricting each update to a damped refinement. LoopUS realizes this via a selective gate that interpolates proposed updates with the previous state. This dampens latent-space displacement and steers the trajectory toward regions increasingly favoring the correct answer.
Figure~\ref{fig:representation_learning} visualizes this: each gated refinement preserves part of the prior representation while making a directed move toward an answer-supporting latent subspace. Consequently, LoopUS incorporates an input-dependent selective gate after each reasoning iteration. Given the current hidden state $h^{(b)}$, the gate first measures the residual change proposed by the reasoning block and maps it to a positive per-token, per-channel step size:
\begin{equation}
\begin{aligned}
\delta^{(b)} &= \mathcal{M}(h^{(b)}) - h^{(b)},  \qquad
\Delta^{(b)} &= \operatorname{softplus}\!\left( W_{\Delta}\delta^{(b)} + b_{\Delta}\right).
\end{aligned}
\end{equation}
Since the pretrained block $\mathcal{M}$ is highly nonlinear and lacks strict Lipschitz bounds, guaranteeing a formal global contraction is intractable. To effectively mitigate unconstrained drift, LoopUS instead enforces a relaxed, contraction-like iteration. Using a learned channel-wise decay coefficient $A \in \mathbb{R}_{<0}$, it computes a discrete decay factor,
\begin{equation}
\alpha^{(b)} = \exp\!\left(\Delta^{(b)} \odot A\right),
\end{equation}
ensuring $\alpha^{(b)} \in (0,1)$ elementwise, an approach that shares conceptual synergy with input-dependent decay mechanisms in recent sequence models like Mamba~\cite{gu2024mambalineartimesequencemodeling}. The subsequent hidden state is then obtained by interpolating between the proposed update and the prior state:

\begin{equation}
 h^{(b+1)} = \mathcal{G}\!\left(\mathcal{M}, h^{(b)}\right) = \alpha^{(b)} \odot \mathcal{M}(h^{(b)}) + \left(1 - \alpha^{(b)}\right) \odot h^{(b)}
\label{eq:loopus_gate}
\end{equation}
Since $\alpha^{(b)} \in (0,1)$, Equation~\ref{eq:loopus_gate} provides a convex interpolation between the proposed update and the previous state. Although this convex combination does not mathematically guarantee the entire composite operator is a strict contraction, it restricts the maximal stride of each update. Consequently, each iteration acts as a damped relaxation step---analogous to an Euler integration step in a bounded vector field---rather than an extrapolative update that might amplify drift. Specifically, larger values of $\alpha^{(b)}$ weight the new update more heavily, whereas smaller values preserve more of the prior state:

\begin{equation}
h^{(b+1)} - h^{(b)} = \alpha^{(b)} \odot \left(\mathcal{M}(h^{(b)}) - h^{(b)}\right),
\end{equation}

or, expressed in vector form:

\begin{equation}
h^{(b+1)} = h^{(b)} + P^{(b)}\left(\mathcal{M}(h^{(b)}) - h^{(b)}\right),
\qquad
P^{(b)} = \operatorname{Diag}(\alpha^{(b)}).
\end{equation}
Under a continuous-time analogy, this recursion corresponds to a forward Euler step for the state-dependent ordinary differential equation:
\begin{equation}
\dot{h} = P(h)\left(\mathcal{M}(h) - h\right),
\end{equation}
where $P(h)$ acts as a diagonal preconditioner induced by the gate. Because its diagonal entries lie strictly in $(0,1)$, the gate applies a damped step size along each coordinate. The discrete update therefore realizes a diagonally preconditioned, relaxed fixed-point iteration toward $h^{\star} = \mathcal{M}(h^{\star})$, where the data-dependent step sizes serve as an implicit per-coordinate regularizer. This design encourages contraction-like behavior across loop iterations, as empirically confirmed in Section~\ref{sec:dynamics_of_stable_latent_refinement}, enabling the stable reuse of pretrained middle layers without architectural modification.

\paragraph{Adaptive Computation via Early Stopping Mechanism.}
\label{sec:method_confidence_head}
To enable adaptive computation at inference time, LoopUS augments each reasoning step with a confidence-based stopping rule. After the $b$-th refinement step, the confidence head produces a raw logit and its corresponding probability:
\begin{equation}
\tilde{q}^{(b)} = q_{\phi}\!\left(h^{(b)}\right), \qquad
q^{(b)} = \sigma\!\left(\tilde{q}^{(b)}\right).
\end{equation}
The model compares $q^{(b)}$ against a predefined threshold $q_{\mathrm{th}}$, continuing to refine the representation while $q^{(b)} < q_{\mathrm{th}}$ and halting once $q^{(b)} \ge q_{\mathrm{th}}$. This reflects the adaptive-computation principle of \emph{Less is More}~\cite{jolicoeurmartineau2025morerecursivereasoningtiny}: additional loop steps are allocated only when the current latent state lacks sufficient confidence. In this way, pretrained transformer depth is dynamically converted into adaptive TTC.

\paragraph{Random Deep Supervision for Loop Training.}
\label{sec:method_random_deep_supervision}

Backpropagating through all loop steps would tightly couple the fully unrolled graph, rendering training memory-intensive and unstable~\cite{phd/ca/Sutskever13}. Thus, LoopUS employs \textbf{random deep supervision}~\cite{wang2025hierarchicalreasoningmodel}: for each training batch, the model is unrolled for $B$ steps, but gradients are computed only for a uniformly sampled subset of steps $\mathcal{S} \subseteq \{0, \dots, B-1\}$ with size $|\mathcal{S}| = K$. Steps in $\mathcal{S}$ receive normal gradient updates, whereas the intermediate steps are executed without gradient tracking (\texttt{no\_grad}) and detached before the subsequent iteration, effectively blocking gradient flow through unsupervised depths. Coupled with the stabilizing effect of the selective gate, this strategy trains the model to halt robustly at diverse stopping depths while circumventing the prohibitive cost of full BPTT~\cite{Peebles2022DiT}.

\subsection{Training Objective}
\label{sec:method_training_objective}

As illustrated in Figure~\ref{fig:framework}(b), LoopUS is trained by jointly optimizing a next-token prediction loss, a monotonicity loss, and a confidence loss at each sampled depth $b \in \mathcal{S}$.

\paragraph{Overall Objective.}
At a sampled depth $b \in \mathcal{S}$, the total per-step loss is defined as:

\begin{equation}
\mathcal{L}^{(b)} = \mathcal{L}_{\mathrm{LM}}^{(b)} + \mathcal{L}_{\mathrm{mono}}^{(b)} + \mathcal{L}_{Q}^{(b)},  
\end{equation}

where $\mathcal{L}_{\mathrm{LM}}^{(b)}$ and $\mathcal{L}_{\mathrm{mono}}^{(b)}$ optimize latent refinement, and $\mathcal{L}_{Q}^{(b)}$ trains early stopping.

\paragraph{Refinement Losses.}
To optimize latent refinement, we employ an autoregressive cross-entropy loss alongside a monotonicity regularizer. The primary supervision acts on the updated logits:

\begin{equation}
\mathcal{L}_{\mathrm{LM}}^{(b)} = \mathrm{CE}\!\left(\mathcal{D}(h^{(b)}), x_{2:T}\right),
\end{equation}

which directly drives the refined latent state to deliver better predictive distributions. To prevent detrimental updates, we evaluate the pre-update state and systematically penalize predictive regressions:

\begin{equation}
\begin{aligned}
\mathcal{L}_{\mathrm{mono}}^{(b)} &= \operatorname{SiLU}\!\left(\mathcal{L}_{\mathrm{LM}}^{(b)} - \mathcal{L}_{\mathrm{LM}}^{(b-1)}\right).
\end{aligned}
\end{equation}
This monotonicity term penalizes updates that degrade the subsequent prediction loss, while remaining negligible for updates that preserve or enhance predictive quality. We adopt the SiLU activation~\cite{elfwing2018sigmoid} because, unlike ReLU~\cite{nair2010rectified} or SELU~\cite{klambauer2017self}, it yields small negative values for minor improvements while asymptoting to zero for large negative arguments. This softly rewards beneficial refinements, encourages the loop to progress via small, stable updates, and stabilizes training without enabling the monotonicity penalty to dominate the primary objective $\mathcal{L}_{\mathrm{LM}}$. Effectively, the monotonicity term enforces a gradual decay in the task-aligned surrogate error across successive loop iterations.

\begin{figure}[htbp]
  \centering
  \begin{lstlisting}[language=python,basicstyle=\ttfamily\small]
h = Encoder(x)
sampled = RandomSampler(B, K)
for b in range(B):
  # Supervised step
  if b in sampled:
    h_prev = h
    h_prop = Reasoner(h)
    h = Gate(h_prev, h_prop)
    q_logit = ConfidenceHead(h)
    y_hat = Decoder(h)
    y_prev = Decoder(h_prev)
    loss = CE(y_hat, y_true)  # L_LM
    loss += SiLU(CE(y_hat, y_true)
                 - CE(y_prev, y_true))  # L_Mono
    loss += BCEWithLogits(q_logit,
                          (y_hat == y_true))  # L_Q
    loss.backward()
    opt.step()
    opt.zero_grad()
    h = h.detach()
  else:  # Unsupervised step
    with no_grad():
      h = Gate(h, Reasoner(h))
    h = h.detach()
  \end{lstlisting}
  \vspace{0.5\baselineskip}
  \caption{Pseudocode of LoopUS.}
  \label{fig:pseudocode}
\end{figure}

\paragraph{Confidence Loss.}
To train adaptive stopping, we supervise the post-update confidence logit $\tilde{q}^{(b+1)}$ with per-sample token accuracy,
\begin{equation}
\begin{aligned}
\mathcal{L}_{Q}^{(b)} &= \mathrm{BCEWithLogits}(\tilde{q}^{(b+1)}, q_{\mathrm{target}}^{(b)}), \\
q_{\mathrm{target}}^{(b)} &= \frac{1}{T_{\mathrm{valid}}} \sum_{j=1}^{T-1}
\mathbf{1}\!\left[\hat{x}_{j}^{(b+1)} = x_{j+1}\right],
\end{aligned}
\end{equation}
This formulation yields a lightweight stopping criterion that requires only a single scalar prediction per step, avoiding the extra statistics required by convergence-based~\cite{gladstone2026energybased} or cumulative distribution function (CDF)-based adaptive rules~\cite{zhu2025scaling}. Together, these terms train LoopUS to make each loop step predictive, avoid regressive updates, and estimate whether further computation is unnecessary.
\section{Empirical Validation}

\subsection{Evaluation Protocol}

We evaluate LoopUS across five pretrained backbones spanning model families and scales: Qwen3-1.7B, Qwen3-4B, and Qwen3-8B~\cite{yang2025qwen3technicalreport}, using cloud NVIDIA L40S, RTX PRO 6000, and RTX PRO 6000 GPUs, respectively; TinyLlama~\cite{zhang2024tinyllama}, using NVIDIA L40S GPUs; and Phi-4~\cite{abdin2024phi4technicalreport}, using NVIDIA H200 GPUs. Unless otherwise stated, models are trained on FineWeb-Edu~\citep{penedo2024the} with 3B tokens, a context length of 1024, the AdamW optimizer, a cosine learning-rate schedule, bf16 mixed precision, and the default LoopUS setting of $B=20$ total loop steps with $K=5$ supervised depths per batch. Models are evaluated with \texttt{lm-evaluation-harness}~\cite{eval-harness}. We report perplexity on WikiText~\cite{merity2017pointer} and Lambada~\cite{paperno-etal-2016-lambada}, and accuracy on MMLU~\cite{hendrycks2021measuringmassivemultitasklanguage}, HellaSwag (HS)~\cite{zellers2019hellaswagmachinereallyfinish}, ARC-Easy (ARC-E), ARC-Challenge (ARC-C)~\cite{allenai:arc}, PIQA~\cite{bisk2019piqareasoningphysicalcommonsense}, WinoGrande (WG)~\cite{sakaguchi2019winograndeadversarialwinogradschema}, and OpenBookQA (OBQA)~\cite{mihaylov2018suitarmorconductelectricity}. Unless otherwise noted, inference uses a maximum recursion budget of 8 with confidence-based stopping and KV caching. Full details are provided in Appendix~\ref{appendix:experimental_details}.

\subsection{Backbone-Level Evaluation across Model Scales}

\begin{table}[ht!]
\caption{\textbf{LoopUS improves pretrained backbones across scales.} Results on language modeling and downstream benchmarks. \texttt{ppl} denotes perplexity (lower is better), and \texttt{acc} denotes accuracy (higher is better). \textbf{AVG} is the mean over the seven \texttt{acc} benchmarks, and $\Delta$ denotes the change in \textbf{AVG} from the original backbone (w/o LoopUS) to the adapted checkpoint (w/ LoopUS). \textbf{Bold} highlights the better result between the two variants of each backbone. All models are evaluated zero-shot.}
\label{tab:main_results}
\centering
\resizebox{\textwidth}{!}{%
 \begin{tabular}{lcccccccccccc}
\toprule
\multirow{2}{*}{Model} & \multirow{2}{*}{Setting} & Wiki & LAMBADA & MMLU & HS & ARC-E & ARC-C & PIQA & WG & OBQA & \multirow{2}{*}{AVG} & \multirow{2}{*}{$\Delta$} \\
\cmidrule(lr){3-4} \cmidrule(lr){5-11}
 &  & \multicolumn{2}{c}{\texttt{ppl} $\downarrow$} & \multicolumn{7}{c}{\texttt{acc} $\uparrow$} &  &  \\
\midrule
\multirow{2}{*}{Qwen 1.7B} & w/o LoopUS & 21 & 12.21 & 55.4 & 46.2 & 72.5 & 40.2 & 72.2 & 61.3 & 28 & 53.7 & -- \\
 & w/ LoopUS & \textbf{16.9} & \textbf{7.43} & \textbf{56.6} & \textbf{46.3} & \textbf{74.9} & \textbf{43.1} & \textbf{73.3} & \textbf{63.0} & \textbf{29.6} & \textbf{55.3} & \textbf{+1.6} \\
\midrule
\multirow{2}{*}{Qwen 4B} & w/o LoopUS & 16.4 & 7.29 & \textbf{68.3} & \textbf{52.1} & 80.2 & 50.4 & 75.0 & 66.5 & 29.4 & 60.3 & -- \\
 & w/ LoopUS & \textbf{13.9} & \textbf{5.33} & 67.7 & 51.4 & \textbf{81.3} & \textbf{54.0} & \textbf{76.8} & \textbf{68.9} & \textbf{34.4} & \textbf{62.1} & \textbf{+1.8} \\
\midrule
\multirow{2}{*}{Qwen 8B} & w/o LoopUS & 12.2 & 4.58 & \textbf{72.8} & \textbf{57.2} & 81.5 & 55.4 & 76.3 & 67.9 & 31.6 & 63.2 & -- \\
 & w/ LoopUS & \textbf{10.3} & \textbf{4.32} & 71.5 & 56.0 & \textbf{83.9} & \textbf{58.1} & \textbf{78.9} & \textbf{72.4} & \textbf{37.0} & \textbf{65.4} & \textbf{+2.2} \\
\midrule
\multirow{2}{*}{Phi-4 14B} & w/o LoopUS & 9.59 & 4.03 & 76.9 & \textbf{63.1} & 81.3 & 55.8 & 80.7 & 77.0 & 34.0 & 67.0 & -- \\
 & w/ LoopUS & \textbf{7.75} & \textbf{3.49} & \textbf{77.5} & 60.58 & \textbf{83.5} & \textbf{57.7} & \textbf{81.8} & \textbf{77.5} & \textbf{41.8} & \textbf{68.6} & \textbf{+1.7} \\
\bottomrule
\end{tabular}%
}
\end{table}

LoopUS reuses pretrained computation by partitioning the backbone into encoder, reasoning, and decoder blocks while preserving the external decoding interface. Table~\ref{tab:main_results} shows that this recasting yields consistent gains across models, reducing WikiText and LAMBADA perplexities and improving average downstream accuracy by +1.6 to +2.2 points, with the clearest gains on ARC-C and OBQA.

The effect is task-dependent: MMLU and HS remain close to the base models, whereas ARC-C, PIQA, WG, and OBQA improve more consistently. This pattern suggests that LoopUS is most useful when extra latent computation can refine a decision process, and less so when performance depends more on broad knowledge retrieval or on already strong single-pass predictions. The same reasoning-oriented trend holds across model scales, indicating that architectural recasting provides a stable post-training modification rather than a task-specific patch.

\subsection{Comparison with Prior Methods under Limited Training Budgets}

\begin{table*}[ht!]
\caption{\textbf{LoopUS shows adaptation efficiency under a smaller training-token budget.} All methods adapt a TinyLlama-based backbone; w/o and w/ LoopUS denote the checkpoint before and after adaptation, respectively. AVG is the unweighted mean over the six tasks, and $\Delta$ reports the change in AVG from Original to Adapted. Results for prior methods are taken from the corresponding papers.}

\label{tab:published_looped_comparison}
\centering
\small
\resizebox{\textwidth}{!}{%
\begin{tabular}{cccccccccccc}
\toprule
\multirow{3}{*}{Method} & \multirow{3}{*}{\makecell[c]{Base\\Model}} & \multirow{3}{*}{\makecell[c]{Train\\Tokens}} & \multirow{3}{*}{Setting} & \multicolumn{6}{c}{Task} & \multirow{3}{*}{AVG} & \multirow{3}{*}{$\Delta$} \\
\cmidrule(lr){5-10}
 &  &  &  & ARC-E & ARC-C & HS & WG & PIQA & OBQA &  &  \\
\cmidrule(lr){5-10}
 &  &  &  & \texttt{acc\_n $\uparrow$} & \texttt{acc\_n $\uparrow$} & \texttt{acc\_n $\uparrow$} & \texttt{acc $\uparrow$} & \texttt{acc\_n $\uparrow$} & \texttt{acc\_n $\uparrow$} &  &  \\
\midrule
\multirow{2}{*}{Ours} & \multirow{2}{*}{\href{https://huggingface.co/TinyLlama/TinyLlama_v1.1}{\makecell[c]{TinyLlama\\1.1B}}} & \multirow{2}{*}{3B} & Original & 47.1 & 25.1 & 42.2 & 53.4 & 66.8 & 24.2 & 43.1 & -- \\
 &  &  & Adapted & \textbf{53.0} & \textbf{29.6} & \textbf{55.5} & \textbf{57.9} & \textbf{69.8} & \textbf{30.6} & \textbf{49.4} & \textbf{+6.3} \\
\midrule
\multirow{2}{*}{\makecell[c]{McLeish et al.\\(\citeyear{mcleish2025retrofitted})}} & \multirow{2}{*}{\href{https://huggingface.co/TinyLlama/TinyLlama-1.1B-intermediate-step-1431k-3T}{\makecell[c]{TinyLlama\\1.1B-3T}}} & \multirow{2}{*}{52B} & Original & 55.7 & 31.0 & \textbf{59.1} & \textbf{58.9} & \textbf{73.0} & \textbf{35.0} & \textbf{52.1} & -- \\
 &  &  & Adapted & \textbf{58.6} & \textbf{35.6} & 45.1 & 57.6 & 66.4 & 32.2 & 49.3 & -2.9 \\
\midrule
\multirow{2}{*}{\makecell[c]{Bae et al.\\(\citeyear{bae2025relaxedrecursive})}} & \multirow{2}{*}{\href{https://huggingface.co/TinyLlama/TinyLlama_v1.1}{\makecell[c]{TinyLlama\\1.1B}}} & \multirow{2}{*}{60B} & Original & 44.7 & 23.2 & 42.2 & 53.4 & 66.8 & 29.2 & 43.3 & -- \\
 &  &  & Adapted & \textbf{49.9} & \textbf{26.2} & \textbf{48.8} & \textbf{54.1} & \textbf{68.6} & \textbf{32.8} & \textbf{46.7} & \textbf{+3.5} \\
\bottomrule
\end{tabular}%
}
\end{table*}

LoopUS is designed to keep loop training stable and adaptation-efficient through selective gating and sparse supervision across depths. Table~\ref{tab:published_looped_comparison} shows the practical effect of this design choice on a shared six-task reasoning suite. Since prior results are drawn from the corresponding papers, we treat this comparison as an adaptation-efficiency reference rather than a fully controlled head-to-head benchmark. In this comparison, LoopUS achieves the largest average gain ($\Delta{=}+6.3$), compared with $\Delta{=}-2.9$ for \citet{mcleish2025retrofitted} and $\Delta{=}+3.5$ for \citet{bae2025relaxedrecursive}, while using fewer additional training tokens. These results suggest that LoopUS improves adaptation efficiency not simply by adding recurrence but by preserving and reusing pretrained computation through decomposition, selective gating, and random deep supervision.

\subsection{Inference-Time Recursion-Depth Analysis}
\label{sec:recursion_depth_diagnostics}

\begin{figure}[ht!]
\centering
\includegraphics[width=0.95\textwidth]{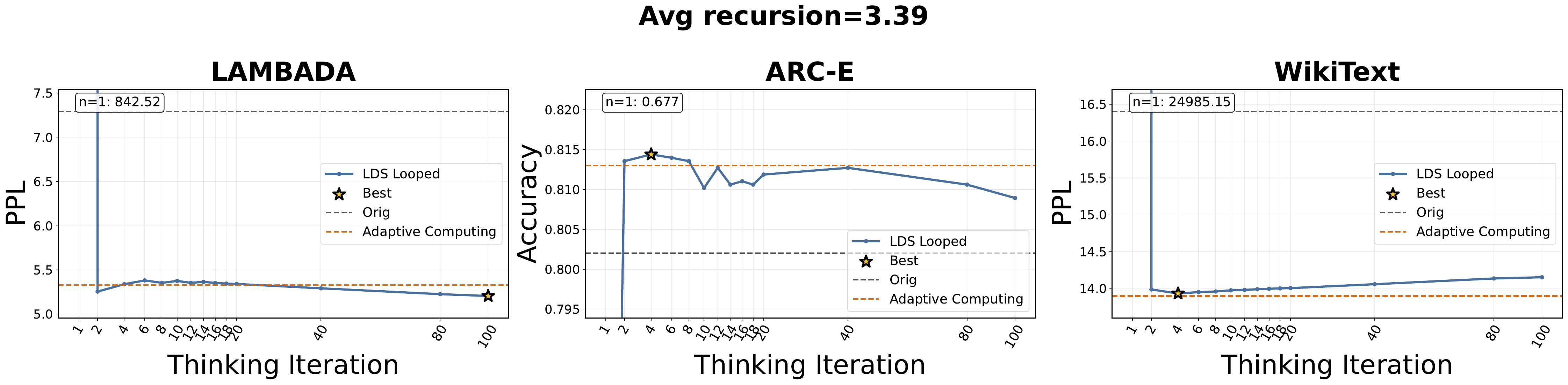}
\caption{\textbf{Test-Time Scaling of LoopUS.} On Qwen3-4B, benchmark performance is plotted against the number of latent reasoning iterations used at inference time. The dashed gray line denotes the original backbone, the dashed orange line denotes the trained LoopUS checkpoint, and the star marks the best observed depth for each task.}
\label{fig:qwen4b_recursion_task_trends}
\end{figure}

LoopUS uses a confidence-based stopping rule to allocate TTC adaptively. Figure~\ref{fig:qwen4b_recursion_task_trends} shows that most of the benefit is obtained within only a few iterations, after which additional recursion yields diminishing returns while remaining stable rather than diverging. This stability extends well beyond the training regime: the checkpoint continues to behave robustly even at unseen recursion depths such as 40, 80, and 100. With adaptive stopping enabled, the same checkpoint halts after 3.39 iterations on average out of a maximum budget of 8, yet remains close to the best observed performance. These results suggest that the confidence head does not merely stop early; it learns to identify an effective stopping point quickly and allocate extra refinement only when it is useful.

\subsection{Dynamics of Stable Latent Refinement}
\label{sec:dynamics_of_stable_latent_refinement}
\begin{figure}[ht!]
\centering
\includegraphics[width=\textwidth]{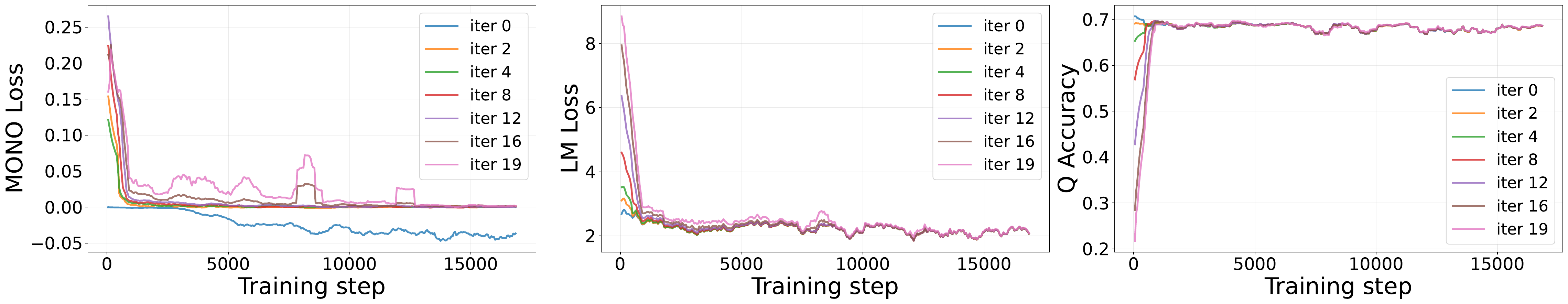}
\caption{\textbf{Training organizes the loop into a stable refinement process.} We plot the step-wise monotonicity loss, next-token prediction loss, and confidence-head Accuracy for loop indices $\{0, 2, 4, 8, 12, 16, 19\}$.}
\label{fig:qwen8b_metrics} 
\end{figure}

LoopUS trains each loop step as a damped corrective update through selective gating and a monotonicity-aware objective. This is consistent with recent energy-based views of autoregressive modeling, where extra latent computation acts as iterative refinement toward more compatible states~\citep{blondel2026autoregressive,gladstone2026energybased}. Figure~\ref{fig:qwen8b_metrics} shows this behavior emerging during training: the monotonicity loss decreases toward zero across loop positions, while the next-token prediction loss and confidence loss remain well-behaved across shallow and deep unrolls. Training therefore encourages each transition to be a small, stable corrective edit rather than an unstable depth expansion.

\begin{figure}[ht!]
\centering
\includegraphics[width=\textwidth]{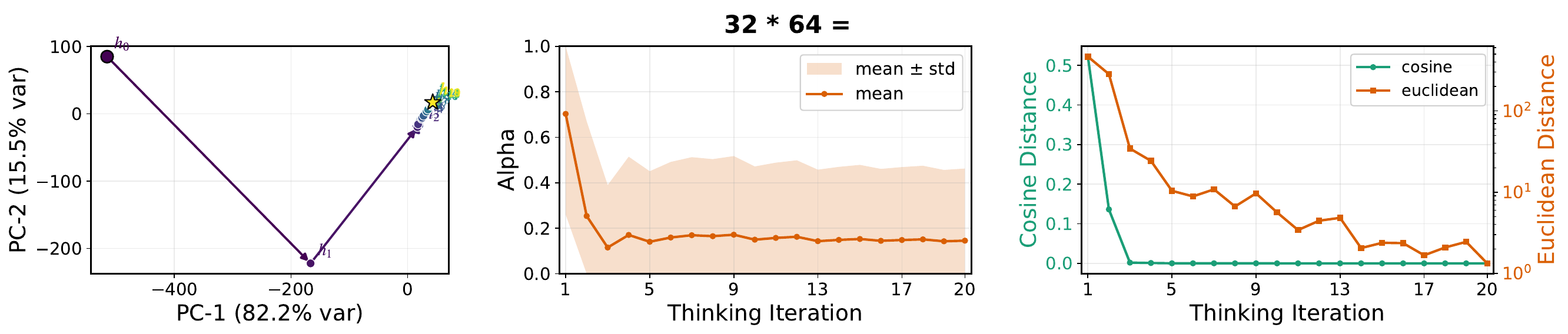}
\caption{\textbf{The learned loop induces convergent trajectories.} The largest latent-space movement occurs in the earliest iterations, after which the step-to-step distance contracts, indicating that the latent trajectory approaches a fixed point rather than diverging.}
\label{fig:loopus_distance_qwen4b}
\end{figure}

\begin{figure}[ht!]
\centering
\includegraphics[width=0.7\textwidth]{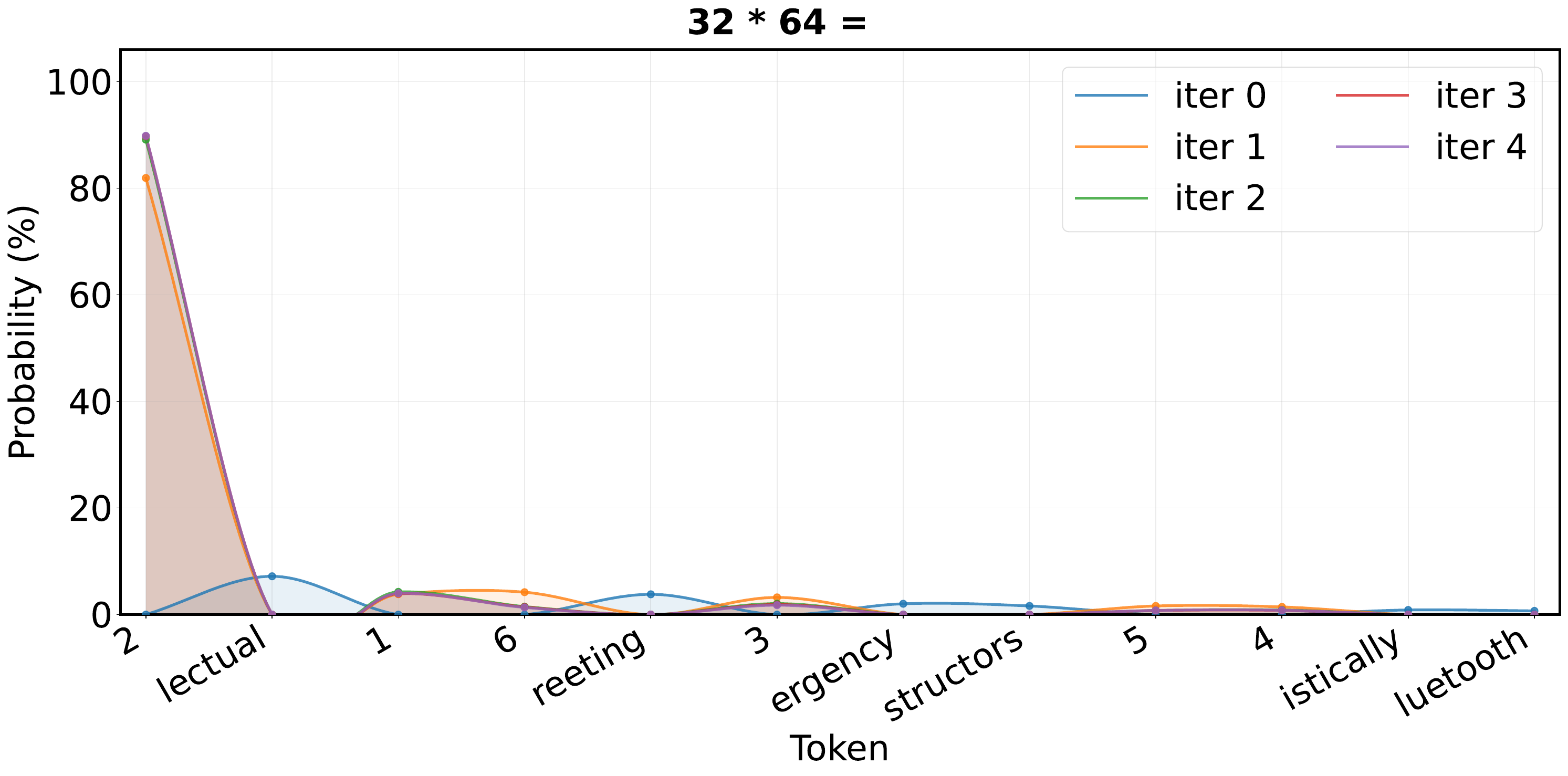}
\caption{\textbf{Loop updates translate into token-level predictive refinement.} Across iterations, probability mass shifts across candidate tokens, showing how latent updates refine the next-token prediction.}
\label{fig:loopus_next_token_distribution_qwen3_4b}
\end{figure}

Figures~\ref{fig:loopus_distance_qwen4b} and~\ref{fig:loopus_next_token_distribution_qwen3_4b} show the same Qwen3-4B example for the prompt ``32 * 64 ='' from latent- and token-space perspectives, respectively. The latent trajectory makes its largest move in the first few iterations and then contracts, indicating convergence toward a stable answer region. Consistently, the correct next token ``2'' rises from $2.17 \times 10^{-5}\%$ at iteration 0 to $81.9\%$ after one refinement step and to about $89.8\%$ by iteration 4, while the remaining candidates lose most of their mass early on. Together with Figure~\ref{fig:qwen8b_metrics}, these results suggest that LoopUS uses a large initial corrective update followed by smaller, convergent refinements that sharpen the final prediction.

\subsection{Component Ablation Study}

\begin{figure}[ht!]
\centering
\includegraphics[width=\textwidth]{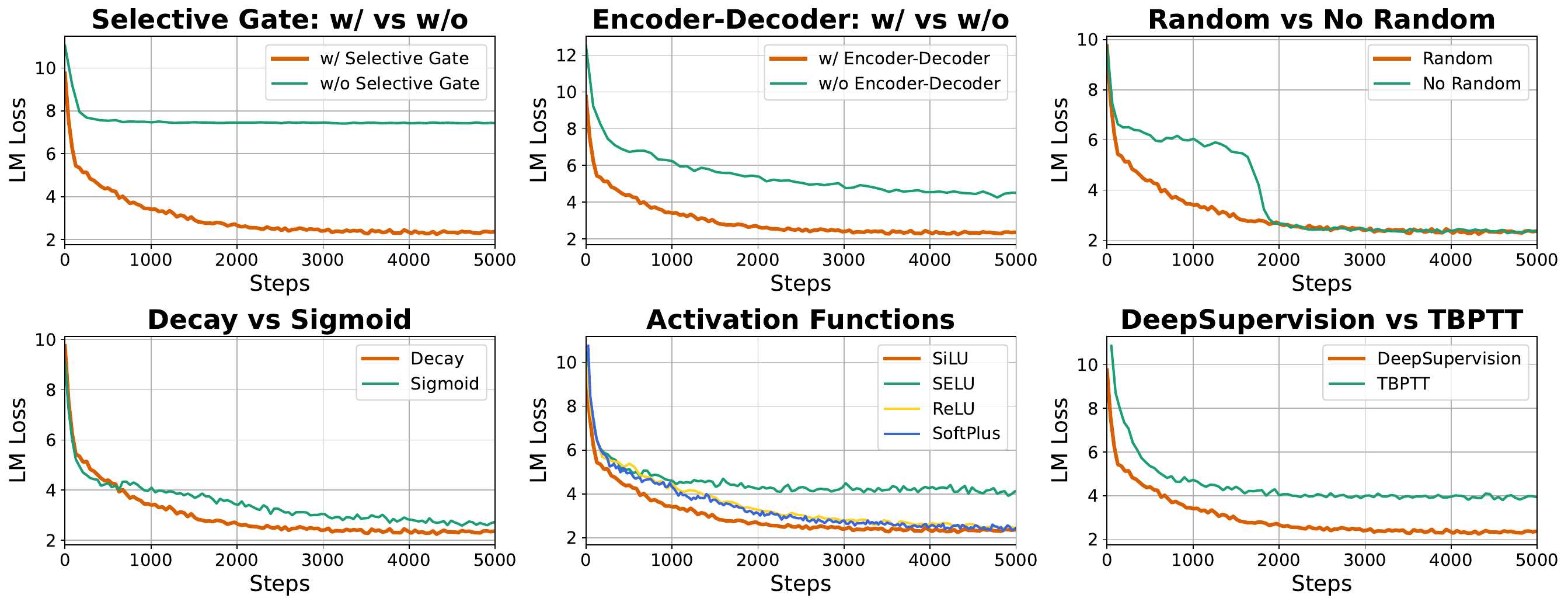}
\caption{\textbf{Ablation study of LoopUS components.} We report average $\mathcal{L}_{\mathrm{LM}}$ over 20 runs after (a) removing the selective gate, (b) removing the encoder-decoder decomposition, (c) training without random deep supervision, (d) replacing the decay gate with sigmoid gating, (e) changing the monotonicity-loss activation among ReLU, SiLU, SELU, and SoftPlus, and (f) comparing the standard LoopUS training recipe against TBPTT.}
\label{fig:ablation}
\end{figure}

Figure~\ref{fig:ablation} analyzes how $\mathcal{L}_{\mathrm{LM}}$ changes when key components of LoopUS are removed or replaced. \textbf{(a)} Removing the selective gate causes convergence to a higher $\mathcal{L}_{\mathrm{LM}}$ because it eliminates the damped interpolation that preserves the previous hidden state, thereby weakening drift-controlled latent refinement. \textbf{(b)} Removing the encoder--decoder decomposition also leads to a higher final $\mathcal{L}_{\mathrm{LM}}$. This demonstrates that without an explicit separation between representation extraction, latent refinement, and output decoding, the loop fails to preserve the pretrained latent workspace. Instead, it relearns a less stable recurrent trajectory that converges to a worse optimum. \textbf{(c)} Training without random deep supervision destabilizes optimization and slows convergence, even if the final $\mathcal{L}_{\mathrm{LM}}$ remains similar. This highlights that sparse supervision across depths is critical for efficiently training long loops in practice. \textbf{(d)} Replacing the decay gate with sigmoid gating also makes optimization less stable and yields a higher final $\mathcal{L}_{\mathrm{LM}}$, indicating that the decay-style gate better supports stable long-loop training. \textbf{(e)} Among the tested activation functions for the monotonicity term, SiLU~\cite{elfwing2018sigmoid} provides the most reliable optimization behavior. ReLU~\cite{nair2010rectified}, SELU~\cite{klambauer2017self}, and SoftPlus~\cite{NIPS2000_44968aec} each lead to less stable or less favorable trajectories, which supports the design choice used in the main LoopUS recipe. \textbf{(f)} TBPTT~\cite{818041} incurs higher computational cost while plateauing at a substantially higher $\mathcal{L}_{\mathrm{LM}}$ than the standard LoopUS training recipe, indicating lower efficiency and worse performance in this setting.
\section{Conclusion}
This paper presents Looped Depth Up-Scaling (LoopUS), a post-training framework that recasts a pretrained LLM into a looped latent-refinement model through encoder--reasoning--decoder decomposition, a selective gate, random deep supervision with stepwise detachment, and a lightweight confidence head for adaptive stopping. Across diverse model scales, LoopUS improves pretrained backbones while preserving standard interfaces, yielding enhanced reasoning performance, consistent perplexity reductions, and high adaptation efficiency under limited training budgets. Our analyses suggest that these gains arise from controlled latent refinement rather than uncontrolled depth expansion. Training remains well behaved across loop depths, with hidden-state trajectories contracting through diminishing corrections, token distributions becoming sharper, and adaptive halting allocating computation where it is most useful. Ablations further confirm that the selective gate, architectural decomposition, and random deep supervision are central to making long latent loops effective and trainable. Overall, latent looping provides a practical way to turn pretrained transformer depth into an adaptive allocation of test-time compute and stronger task-aligned inference.

\bibliographystyle{unsrtnat}
\bibliography{references}

@inproceedings{lahoti2026mamba,
title={Mamba-3: Improved Sequence Modeling using State Space Principles},
author={Aakash Lahoti and Kevin Li and Berlin Chen and Caitlin Wang and Aviv Bick and J Zico Kolter and Tri Dao and Albert Gu},
booktitle={The Fourteenth International Conference on Learning Representations},
year={2026},
url={https://openreview.net/forum?id=HwCvaJOiCj}
}

@inproceedings{snell2025scaling,
title={Scaling {LLM} Test-Time Compute Optimally Can be More Effective than Scaling Parameters for Reasoning},
author={Charlie Victor Snell and Jaehoon Lee and Kelvin Xu and Aviral Kumar},
booktitle={The Thirteenth International Conference on Learning Representations},
year={2025},
url={https://openreview.net/forum?id=4FWAwZtd2n}
}

@inproceedings{
wu2025inference,
title={Inference Scaling Laws: An Empirical Analysis of Compute-Optimal Inference for {LLM} Problem-Solving},
author={Yangzhen Wu and Zhiqing Sun and Shanda Li and Sean Welleck and Yiming Yang},
booktitle={The Thirteenth International Conference on Learning Representations},
year={2025},
url={https://openreview.net/forum?id=VNckp7JEHn}
}

@inproceedings{wei2022chain,
title={Chain of Thought Prompting Elicits Reasoning in Large Language Models},
author={Jason Wei and Xuezhi Wang and Dale Schuurmans and Maarten Bosma and brian ichter and Fei Xia and Ed H. Chi and Quoc V Le and Denny Zhou},
booktitle={Advances in Neural Information Processing Systems},
editor={Alice H. Oh and Alekh Agarwal and Danielle Belgrave and Kyunghyun Cho},
year={2022},
url={https://openreview.net/forum?id=_VjQlMeSB_J}
}

@inproceedings{gladstone2026energybased,
title={Energy-Based Transformers are Scalable Learners and Thinkers},
author={Alexi Gladstone and Ganesh Nanduru and Md Mofijul Islam and Peixuan Han and Hyeonjeong Ha and Aman Chadha and Yilun Du and Heng Ji and Jundong Li and Tariq Iqbal},
booktitle={The Fourteenth International Conference on Learning Representations},
year={2026},
url={https://openreview.net/forum?id=ZBj3Qp1bYg}
}

@misc{blondel2026autoregressive,
      title={Autoregressive Language Models are Secretly Energy-Based Models: Insights into the Lookahead Capabilities of Next-Token Prediction},
      author={Mathieu Blondel and Michael E. Sander and Germain Vivier-Ardisson and Tianlin Liu and Vincent Roulet},
      year={2026},
      eprint={2512.15605},
      archivePrefix={arXiv},
      primaryClass={cs.LG},
      url={https://arxiv.org/abs/2512.15605},
}

@misc{wang2025hierarchicalreasoningmodel,
      title={Hierarchical Reasoning Model}, 
      author={Guan Wang and Jin Li and Yuhao Sun and Xing Chen and Changling Liu and Yue Wu and Meng Lu and Sen Song and Yasin Abbasi Yadkori},
      year={2025},
      eprint={2506.21734},
      archivePrefix={arXiv},
      primaryClass={cs.AI},
      url={https://arxiv.org/abs/2506.21734}, 
}

@inproceedings{nie2025large,
title={Large Language Diffusion Models},
author={Shen Nie and Fengqi Zhu and Zebin You and Xiaolu Zhang and Jingyang Ou and Jun Hu and JUN ZHOU and Yankai Lin and Ji-Rong Wen and Chongxuan Li},
booktitle={The Thirty-ninth Annual Conference on Neural Information Processing Systems},
year={2025},
url={https://openreview.net/forum?id=KnqiC0znVF}
}

@misc{jolicoeurmartineau2025morerecursivereasoningtiny,
      title={Less is More: Recursive Reasoning with Tiny Networks}, 
      author={Alexia Jolicoeur-Martineau},
      year={2025},
      eprint={2510.04871},
      archivePrefix={arXiv},
      primaryClass={cs.LG},
      url={https://arxiv.org/abs/2510.04871}, 
}

@misc{zhu2025surveylatentreasoning,
      title={A Survey on Latent Reasoning}, 
      author={Rui-Jie Zhu and Tianhao Peng and Tianhao Cheng and Xingwei Qu and Jinfa Huang and Dawei Zhu and Hao Wang and Kaiwen Xue and Xuanliang Zhang and Yong Shan and Tianle Cai and Taylor Kergan and Assel Kembay and Andrew Smith and Chenghua Lin and Binh Nguyen and Yuqi Pan and Yuhong Chou and Zefan Cai and Zhenhe Wu and Yongchi Zhao and Tianyu Liu and Jian Yang and Wangchunshu Zhou and Chujie Zheng and Chongxuan Li and Yuyin Zhou and Zhoujun Li and Zhaoxiang Zhang and Jiaheng Liu and Ge Zhang and Wenhao Huang and Jason Eshraghian},
      year={2025},
      eprint={2507.06203},
      archivePrefix={arXiv},
      primaryClass={cs.CL},
      url={https://arxiv.org/abs/2507.06203}, 
}

@misc{zhou2026dllmsimplediffusionlanguage,
      title={dLLM: Simple Diffusion Language Modeling}, 
      author={Zhanhui Zhou and Lingjie Chen and Hanghang Tong and Dawn Song},
      year={2026},
      eprint={2602.22661},
      archivePrefix={arXiv},
      primaryClass={cs.CL},
      url={https://arxiv.org/abs/2602.22661}, 
}

@inproceedings{beck2024xlstm,
title={x{LSTM}: Extended Long Short-Term Memory},
author={Maximilian Beck and Korbinian P{\"o}ppel and Markus Spanring and Andreas Auer and Oleksandra Prudnikova and Michael K Kopp and G{\"u}nter Klambauer and Johannes Brandstetter and Sepp Hochreiter},
booktitle={The Thirty-eighth Annual Conference on Neural Information Processing Systems},
year={2024},
url={https://openreview.net/forum?id=ARAxPPIAhq}
}

@misc{gu2024mambalineartimesequencemodeling,
      title={Mamba: Linear-Time Sequence Modeling with Selective State Spaces}, 
      author={Albert Gu and Tri Dao},
      year={2024},
      eprint={2312.00752},
      archivePrefix={arXiv},
      primaryClass={cs.LG},
      url={https://arxiv.org/abs/2312.00752}, 
}

@inproceedings{song2021denoising,
title={Denoising Diffusion Implicit Models},
author={Jiaming Song and Chenlin Meng and Stefano Ermon},
booktitle={International Conference on Learning Representations},
year={2021},
url={https://openreview.net/forum?id=St1giarCHLP}
}

@inproceedings{gu2022efficiently,
title={Efficiently Modeling Long Sequences with Structured State Spaces},
author={Albert Gu and Karan Goel and Christopher Re},
booktitle={International Conference on Learning Representations},
year={2022},
url={https://openreview.net/forum?id=uYLFoz1vlAC}
}

@article{hochreiter1997long,
  title={Long short-term memory},
  author={Hochreiter, Sepp and Schmidhuber, J{\"u}rgen},
  journal={Neural computation},
  volume={9},
  number={8},
  pages={1735--1780},
  year={1997},
  publisher={MIT press}
}

@inproceedings{cho2014learning,
    title = "Learning Phrase Representations using {RNN} Encoder{--}Decoder for Statistical Machine Translation",
    author = {Cho, Kyunghyun  and
      van Merri{\"e}nboer, Bart  and
      Gulcehre, Caglar  and
      Bahdanau, Dzmitry  and
      Bougares, Fethi  and
      Schwenk, Holger  and
      Bengio, Yoshua},
    editor = "Moschitti, Alessandro  and
      Pang, Bo  and
      Daelemans, Walter",
    booktitle = "Proceedings of the 2014 Conference on Empirical Methods in Natural Language Processing ({EMNLP})",
    month = oct,
    year = "2014",
    address = "Doha, Qatar",
    publisher = "Association for Computational Linguistics",
    url = "https://aclanthology.org/D14-1179/",
    doi = "10.3115/v1/D14-1179",
    pages = "1724--1734"
}

@article{behrouz2024titans,
  title={Titans: Learning to memorize at test time},
  author={Behrouz, Ali and Zhong, Peilin and Mirrokni, Vahab},
  journal={arXiv preprint arXiv:2501.00663},
  year={2024}
}

@article{srivastava2015training,
      title={Training Very Deep Networks},
      author={Srivastava, Rupesh Kumar and Greff, Klaus and Schmidhuber, J{"u}rgen},
      journal={Advances in Neural Information Processing Systems Workshop on Deep Learning},
      year={2015},
      url={https://arxiv.org/abs/1507.06228}
}

@misc{nostalgebraist2020logitlens,
      author = {nostalgebraist},
      title = {Interpreting {GPT}: The Logit Lens},
      year = {2020},
      howpublished = {LessWrong},
      url = {https://www.lesswrong.com/posts/AcKRB8wDpdaN6v6ru/interpreting-gpt-the-logit-lens}
}

@inproceedings{voita2019bottomup,
      title = {The Bottom-Up Evolution of Representations in the Transformer: A Study with Machine Translation and Language Modeling Objectives},
      author = {Voita, Elena and Talbot, David and Moiseev, Fedor and Sennrich, Rico and Titov, Ivan},
      booktitle = {Proceedings of the 2019 Conference on Empirical Methods in Natural Language Processing and the 9th International Joint Conference on Natural Language Processing},
      year = {2019},
      pages = {4395--4405},
      url = {https://aclanthology.org/D19-1448/}
}

@ARTICLE{10264112,
  author={Sim, Sunghyun and Kim, Dohee and Bae, Hyerim},
  journal={IEEE Transactions on Pattern Analysis and Machine Intelligence}, 
  title={Correlation Recurrent Units: A Novel Neural Architecture for Improving the Predictive Performance of Time-Series Data}, 
  year={2023},
  volume={45},
  number={12},
  pages={14266-14283},
  doi={10.1109/TPAMI.2023.3319557}}

@inproceedings{FrozenintheMiddle,
author = {Tikhonov, Pavel and Ilvovsky, Dmitry},
title = {Frozen in the Middle: Hidden States Remain Unchanged Across Intermediate Layers of Language Models},
year = {2025},
isbn = {9798400720406},
publisher = {Association for Computing Machinery},
address = {New York, NY, USA},
url = {https://doi.org/10.1145/3746252.3760890},
doi = {10.1145/3746252.3760890},
booktitle = {Proceedings of the 34th ACM International Conference on Information and Knowledge Management},
pages = {5289–5293},
numpages = {5},
location = {Seoul, Republic of Korea},
series = {CIKM '25}
}

@misc{anthropic2024mappingmind,
      author = {{Anthropic}},
      title = {Mapping the Mind of a Large Language Model},
      year = {2024},
      url = {https://www.anthropic.com/research/mapping-mind-language-model}
}

@misc{anthropic2025biology,
      author = {{Anthropic}},
      title = {On the Biology of a Large Language Model},
      year = {2025},
      url = {https://transformer-circuits.pub/2025/attribution-graphs/biology.html}
}

@misc{shibata2026suppressingfinallayerhidden,
      title={Suppressing Final Layer Hidden State Jumps in Transformer Pretraining}, 
      author={Keigo Shibata and Kazuki Yano and Ryosuke Takahashi and Jaesung Lee and Wataru Ikeda and Jun Suzuki},
      year={2026},
      eprint={2601.18302},
      archivePrefix={arXiv},
      primaryClass={cs.CL},
      url={https://arxiv.org/abs/2601.18302}, 
}

@inproceedings{kim-etal-2024-solar,
    title = "{SOLAR} 10.7{B}: Scaling Large Language Models with Simple yet Effective Depth Up-Scaling",
    author = "Kim, Sanghoon  and
      Kim, Dahyun  and
      Park, Chanjun  and
      Lee, Wonsung  and
      Song, Wonho  and
      Kim, Yunsu  and
      Kim, Hyeonwoo  and
      Kim, Yungi  and
      Lee, Hyeonju  and
      Kim, Jihoo  and
      Ahn, Changbae  and
      Yang, Seonghoon  and
      Lee, Sukyung  and
      Park, Hyunbyung  and
      Gim, Gyoungjin  and
      Cha, Mikyoung  and
      Lee, Hwalsuk  and
      Kim, Sunghun",
    editor = "Yang, Yi  and
      Davani, Aida  and
      Sil, Avi  and
      Kumar, Anoop",
    booktitle = "Proceedings of the 2024 Conference of the North American Chapter of the Association for Computational Linguistics: Human Language Technologies (Volume 6: Industry Track)",
    month = jun,
    year = "2024",
    address = "Mexico City, Mexico",
    publisher = "Association for Computational Linguistics",
    url = "https://aclanthology.org/2024.naacl-industry.3/",
    doi = "10.18653/v1/2024.naacl-industry.3",
    pages = "23--35"
}

@misc{dehghani2024universal,
      title={Universal Transformers},
      author={Mostafa Dehghani and Stephan Gouws and Oriol Vinyals and Jakob Uszkoreit and {\L}ukasz Kaiser},
      year={2019},
      eprint={1807.03819},
      archivePrefix={arXiv},
      primaryClass={cs.CL},
      url={https://arxiv.org/abs/1807.03819}
}

@misc{mcleish2025retrofitted,
      title={Teaching Pretrained Language Models to Think Deeper with Retrofitted Recurrence},
      author={Sean McLeish and Ang Li and John Kirchenbauer and Dayal Singh Kalra and Brian R. Bartoldson and Bhavya Kailkhura and Avi Schwarzschild and Jonas Geiping and Tom Goldstein and Micah Goldblum},
      year={2025},
      eprint={2511.07384},
      archivePrefix={arXiv},
      primaryClass={cs.CL},
      url={https://arxiv.org/abs/2511.07384}
}

@misc{geiping2025scalingtesttimecompute,
      title={Scaling up Test-Time Compute with Latent Reasoning: A Recurrent Depth Approach},
      author={Jonas Geiping and Sean McLeish and Neel Jain and John Kirchenbauer and Siddharth Singh and Brian R. Bartoldson and Bhavya Kailkhura and Abhinav Bhatele and Tom Goldstein},
      year={2025},
      eprint={2502.05171},
      archivePrefix={arXiv},
      primaryClass={cs.CL},
      url={https://arxiv.org/abs/2502.05171}
}

@misc{zeng2025ponderlm,
      title={Pretraining Language Models to Ponder in Continuous Space},
      author={Boyi Zeng and Shixiang Song and Siyuan Huang and Yixuan Wang and He Li and Ziwei He and Xinbing Wang and Zhiyu Li and Zhouhan Lin},
      year={2025},
      eprint={2505.20674},
      archivePrefix={arXiv},
      primaryClass={cs.CL},
      url={https://arxiv.org/abs/2505.20674}
}

@misc{fu2025thinkathard,
      title={Think-at-Hard: Teaching Small Language Models to Think on Hard Problems},
      author={Yue Fu and Shruti Rijhwani and Graham Neubig and Yonatan Bisk},
      year={2025},
      eprint={2506.04458},
      archivePrefix={arXiv},
      primaryClass={cs.CL},
      url={https://arxiv.org/abs/2506.04458}
}

@inproceedings{bae2025relaxedrecursive,
title={Relaxed Recursive Transformers: Effective Parameter Sharing with Layer-wise Lo{RA}},
author={Sangmin Bae and Adam Fisch and Hrayr Harutyunyan and Ziwei Ji and Seungyeon Kim and Tal Schuster},
booktitle={The Thirteenth International Conference on Learning Representations},
year={2025},
url={https://openreview.net/forum?id=WwpYSOkkCt}
}

@inproceedings{bae2025mixtureofrecursions,
title={Mixture-of-Recursions: Learning Dynamic Recursive Depths for Adaptive Token-Level Computation},
author={Sangmin Bae and Yujin Kim and Reza Bayat and Sungnyun Kim and Jiyoun Ha and Tal Schuster and Adam Fisch and Hrayr Harutyunyan and Ziwei Ji and Aaron Courville and Se-Young Yun},
booktitle={The Thirty-ninth Annual Conference on Neural Information Processing Systems},
year={2025},
url={https://openreview.net/forum?id=QuqsEIVWIG}
}

@misc{zhu2025ouro,
      title={Ouro: A Latent Reasoning Model with Adaptive Depth via Gated Recurrence},
      author={Rui-Jie Zhu and others},
      year={2025},
      eprint={2507.07919},
      archivePrefix={arXiv},
      primaryClass={cs.CL},
      url={https://arxiv.org/abs/2507.07919}
}

@article{zhu2025scaling,
  title={Scaling latent reasoning via looped language models},
  author={Zhu, Rui-Jie and Wang, Zixuan and Hua, Kai and Zhang, Tianyu and Li, Ziniu and Que, Haoran and Wei, Boyi and Wen, Zixin and Yin, Fan and Xing, He and others},
  journal={arXiv preprint arXiv:2510.25741},
  year={2025}
}

@phdthesis{phd/ca/Sutskever13,
  author = {Sutskever, Ilya},
  ee = {http://hdl.handle.net/1807/36012},
  school = {University of Toronto, Canada},
  title = {Training Recurrent Neural Networks.},
  year = 2013
}

@inproceedings{pascanu2013difficulty,
  title={On the difficulty of training recurrent neural networks},
  author={Pascanu, Razvan and Mikolov, Tomas and Bengio, Yoshua},
  booktitle={International conference on machine learning},
  pages={1310--1318},
  year={2013},
  organization={Pmlr}
}

@inproceedings{yang2025gated,
      title={Gated Delta Networks: Improving Mamba2 with Delta Rule},
      author={Songlin Yang and Jan Kautz and Ali Hatamizadeh},
      booktitle={The Thirteenth International Conference on Learning Representations},
      year={2025},
      url={https://openreview.net/forum?id=r8H7xhYPwz}
}

@article{ng2026rys,
  title   = {LLM Neuroanatomy: How I Topped the LLM Leaderboard Without Changing a Single Weight},
  author  = {Ng, David Noel},
  year    = {2026},
  month   = {March},
  url     = {https://dnhkng.github.io/posts/rys/}
}

@article{klambauer2017self,
  title={Self-normalizing neural networks},
  author={Klambauer, G{\"u}nter and Unterthiner, Thomas and Mayr, Andreas and Hochreiter, Sepp},
  journal={Advances in neural information processing systems},
  volume={30},
  year={2017}
}

@article{elfwing2018sigmoid,
  title={Sigmoid-weighted linear units for neural network function approximation in reinforcement learning},
  author={Elfwing, Stefan and Uchibe, Eiji and Doya, Kenji},
  journal={Neural networks},
  volume={107},
  pages={3--11},
  year={2018},
  publisher={Elsevier}
}

@inproceedings{nair2010rectified,
  title={Rectified linear units improve restricted boltzmann machines},
  author={Nair, Vinod and Hinton, Geoffrey E},
  booktitle={Proceedings of the 27th international conference on machine learning (ICML-10)},
  pages={807--814},
  year={2010}
}

@article{shin2026mi,
  title={Mi: dm 2.0 Korea-centric Bilingual Language Models},
  author={Shin, Donghoon and Lee, Sejung and Bae, Soonmin and Ryu, Hwijung and Ok, Changwon and Jung, Hoyoun and Ji, Hyesung and Lim, Jeehyun and Lee, Jehoon and Han, Ji-Eun and others},
  journal={arXiv preprint arXiv:2601.09066},
  year={2026}
}

@article{Peebles2022DiT,
  title={Scalable Diffusion Models with Transformers},
  author={William Peebles and Saining Xie},
  year={2022},
  journal={arXiv preprint arXiv:2212.09748},
}

@misc{yang2025qwen3technicalreport,
      title={Qwen3 Technical Report}, 
      author={An Yang and Anfeng Li and Baosong Yang and Beichen Zhang and Binyuan Hui and Bo Zheng and Bowen Yu and Chang Gao and Chengen Huang and Chenxu Lv and Chujie Zheng and Dayiheng Liu and Fan Zhou and Fei Huang and Feng Hu and Hao Ge and Haoran Wei and Huan Lin and Jialong Tang and Jian Yang and Jianhong Tu and Jianwei Zhang and Jianxin Yang and Jiaxi Yang and Jing Zhou and Jingren Zhou and Junyang Lin and Kai Dang and Keqin Bao and Kexin Yang and Le Yu and Lianghao Deng and Mei Li and Mingfeng Xue and Mingze Li and Pei Zhang and Peng Wang and Qin Zhu and Rui Men and Ruize Gao and Shixuan Liu and Shuang Luo and Tianhao Li and Tianyi Tang and Wenbiao Yin and Xingzhang Ren and Xinyu Wang and Xinyu Zhang and Xuancheng Ren and Yang Fan and Yang Su and Yichang Zhang and Yinger Zhang and Yu Wan and Yuqiong Liu and Zekun Wang and Zeyu Cui and Zhenru Zhang and Zhipeng Zhou and Zihan Qiu},
      year={2025},
      eprint={2505.09388},
      archivePrefix={arXiv},
      primaryClass={cs.CL},
      url={https://arxiv.org/abs/2505.09388}, 
}

@misc{abdin2024phi4technicalreport,
      title={Phi-4 Technical Report}, 
      author={Marah Abdin and Jyoti Aneja and Harkirat Behl and Sébastien Bubeck and Ronen Eldan and Suriya Gunasekar and Michael Harrison and Russell J. Hewett and Mojan Javaheripi and Piero Kauffmann and James R. Lee and Yin Tat Lee and Yuanzhi Li and Weishung Liu and Caio C. T. Mendes and Anh Nguyen and Eric Price and Gustavo de Rosa and Olli Saarikivi and Adil Salim and Shital Shah and Xin Wang and Rachel Ward and Yue Wu and Dingli Yu and Cyril Zhang and Yi Zhang},
      year={2024},
      eprint={2412.08905},
      archivePrefix={arXiv},
      primaryClass={cs.CL},
      url={https://arxiv.org/abs/2412.08905}, 
}

@inproceedings{penedo2024the,
title={The FineWeb Datasets: Decanting the Web for the Finest Text Data at Scale},
author={Guilherme Penedo and Hynek Kydl{\'\i}{\v{c}}ek and Loubna Ben allal and Anton Lozhkov and Margaret Mitchell and Colin Raffel and Leandro Von Werra and Thomas Wolf},
booktitle={The Thirty-eight Conference on Neural Information Processing Systems Datasets and Benchmarks Track},
year={2024},
url={https://openreview.net/forum?id=n6SCkn2QaG}
}

@misc{eval-harness,
  author       = {Gao, Leo and Tow, Jonathan and Abbasi, Baber and Biderman, Stella and Black, Sid and DiPofi, Anthony and Foster, Charles and Golding, Laurence and Hsu, Jeffrey and Le Noac'h, Alain and Li, Haonan and McDonell, Kyle and Muennighoff, Niklas and Ociepa, Chris and Phang, Jason and Reynolds, Laria and Schoelkopf, Hailey and Skowron, Aviya and Sutawika, Lintang and Tang, Eric and Thite, Anish and Wang, Ben and Wang, Kevin and Zou, Andy},
  title        = {A framework for few-shot language model evaluation},
  month        = 12,
  year         = 2023,
  publisher    = {Zenodo},
  version      = {v0.4.0},
  doi          = {10.5281/zenodo.10256836},
  url          = {https://zenodo.org/records/10256836}
}

@inproceedings{merity2017pointer,
title={Pointer Sentinel Mixture Models},
author={Stephen Merity and Caiming Xiong and James Bradbury and Richard Socher},
booktitle={International Conference on Learning Representations},
year={2017},
url={https://openreview.net/forum?id=Byj72udxe}
}

@inproceedings{paperno-etal-2016-lambada,
    title = "The {LAMBADA} dataset: Word prediction requiring a broad discourse context",
    author = "Paperno, Denis  and
      Kruszewski, Germ{\'a}n  and
      Lazaridou, Angeliki  and
      Pham, Ngoc Quan  and
      Bernardi, Raffaella  and
      Pezzelle, Sandro  and
      Baroni, Marco  and
      Boleda, Gemma  and
      Fern{\'a}ndez, Raquel",
    editor = "Erk, Katrin  and
      Smith, Noah A.",
    booktitle = "Proceedings of the 54th Annual Meeting of the Association for Computational Linguistics (Volume 1: Long Papers)",
    month = aug,
    year = "2016",
    address = "Berlin, Germany",
    publisher = "Association for Computational Linguistics",
    url = "https://aclanthology.org/P16-1144/",
    doi = "10.18653/v1/P16-1144",
    pages = "1525--1534"
}

@misc{zellers2019hellaswagmachinereallyfinish,
      title={HellaSwag: Can a Machine Really Finish Your Sentence?}, 
      author={Rowan Zellers and Ari Holtzman and Yonatan Bisk and Ali Farhadi and Yejin Choi},
      year={2019},
      eprint={1905.07830},
      archivePrefix={arXiv},
      primaryClass={cs.CL},
      url={https://arxiv.org/abs/1905.07830}, 
}

@article{allenai:arc,
      author    = {Peter Clark  and Isaac Cowhey and Oren Etzioni and Tushar Khot and
                    Ashish Sabharwal and Carissa Schoenick and Oyvind Tafjord},
      title     = {Think you have Solved Question Answering? Try ARC, the AI2 Reasoning Challenge},
      journal   = {arXiv:1803.05457v1},
      year      = {2018},
}

@misc{bisk2019piqareasoningphysicalcommonsense,
      title={PIQA: Reasoning about Physical Commonsense in Natural Language}, 
      author={Yonatan Bisk and Rowan Zellers and Ronan Le Bras and Jianfeng Gao and Yejin Choi},
      year={2019},
      eprint={1911.11641},
      archivePrefix={arXiv},
      primaryClass={cs.CL},
      url={https://arxiv.org/abs/1911.11641}, 
}

@misc{sakaguchi2019winograndeadversarialwinogradschema,
      title={WinoGrande: An Adversarial Winograd Schema Challenge at Scale}, 
      author={Keisuke Sakaguchi and Ronan Le Bras and Chandra Bhagavatula and Yejin Choi},
      year={2019},
      eprint={1907.10641},
      archivePrefix={arXiv},
      primaryClass={cs.CL},
      url={https://arxiv.org/abs/1907.10641}, 
}

@misc{mihaylov2018suitarmorconductelectricity,
      title={Can a Suit of Armor Conduct Electricity? A New Dataset for Open Book Question Answering}, 
      author={Todor Mihaylov and Peter Clark and Tushar Khot and Ashish Sabharwal},
      year={2018},
      eprint={1809.02789},
      archivePrefix={arXiv},
      primaryClass={cs.CL},
      url={https://arxiv.org/abs/1809.02789}, 
}

@misc{hendrycks2021measuringmassivemultitasklanguage,
      title={Measuring Massive Multitask Language Understanding}, 
      author={Dan Hendrycks and Collin Burns and Steven Basart and Andy Zou and Mantas Mazeika and Dawn Song and Jacob Steinhardt},
      year={2021},
      eprint={2009.03300},
      archivePrefix={arXiv},
      primaryClass={cs.CY},
      url={https://arxiv.org/abs/2009.03300}, 
}

@inproceedings{dao2024flashattention,
title={FlashAttention-2: Faster Attention with Better Parallelism and Work Partitioning},
author={Tri Dao},
booktitle={The Twelfth International Conference on Learning Representations},
year={2024},
url={https://openreview.net/forum?id=mZn2Xyh9Ec}
}

@misc{chen2026loopbridgeloopedtransformers,
      title={Loop as a Bridge: Can Looped Transformers Truly Link Representation Space and Natural Language Outputs?}, 
      author={Guanxu Chen and Dongrui Liu and Jing Shao},
      year={2026},
      eprint={2601.10242},
      archivePrefix={arXiv},
      primaryClass={cs.CL},
      url={https://arxiv.org/abs/2601.10242}, 
}

@misc{zhang2024tinyllama,
      title={TinyLlama: An Open-Source Small Language Model}, 
      author={Peiyuan Zhang and Guangtao Zeng and Tianduo Wang and Wei Lu},
      year={2024},
      eprint={2401.02385},
      archivePrefix={arXiv},
      primaryClass={cs.CL}
}

@INPROCEEDINGS{818041,
  author={Gers, F.A. and Schmidhuber, J. and Cummins, F.},
  booktitle={1999 Ninth International Conference on Artificial Neural Networks ICANN 99. (Conf. Publ. No. 470)}, 
  title={Learning to forget: continual prediction with LSTM}, 
  year={1999},
  volume={2},
  number={},
  pages={850-855 vol.2},
  doi={10.1049/cp:19991218}}

@article{xu2026looping,
  title={Looping Back to Move Forward: Recursive Transformers for Efficient and Flexible Large Multimodal Models},
  author={Xu, Ruihan and Gao, Yuting and Wang, Lan and Li, Jianing and Chen, Weihao and Guo, Qingpei and Yang, Ming and Zhang, Shiliang},
  journal={arXiv preprint arXiv:2602.09080},
  year={2026}
}

@misc{qwen3.6-27b,
    title  = {{Qwen3.6-27B}: Flagship-Level Coding in a {27B} Dense Model},
    author = {{Qwen Team}},
    month  = {April},
    year   = {2026},
    url    = {https://qwen.ai/blog?id=qwen3.6-27b}
}

@misc{nvidia2025nvidianemotron3efficient,
      title={NVIDIA Nemotron 3: Efficient and Open Intelligence}, 
      author={NVIDIA and : and Aaron Blakeman and Aaron Grattafiori and Aarti Basant and Abhibha Gupta and Abhinav Khattar and Adi Renduchintala and Aditya Vavre and Akanksha Shukla and Akhiad Bercovich and Aleksander Ficek and Aleksandr Shaposhnikov and Alex Kondratenko and Alexander Bukharin and Alexandre Milesi and Ali Taghibakhshi and Alisa Liu and Amelia Barton and Ameya Sunil Mahabaleshwarkar and Amir Klein and Amit Zuker and Amnon Geifman and Amy Shen and Anahita Bhiwandiwalla and Andrew Tao and Anjulie Agrusa and Ankur Verma and Ann Guan and Anubhav Mandarwal and Arham Mehta and Ashwath Aithal and Ashwin Poojary and Asif Ahamed and Asit Mishra and Asma Kuriparambil Thekkumpate and Ayush Dattagupta and Banghua Zhu and Bardiya Sadeghi and Barnaby Simkin and Ben Lanir and Benedikt Schifferer and Besmira Nushi and Bilal Kartal and Bita Darvish Rouhani and Boris Ginsburg and Brandon Norick and Brandon Soubasis and Branislav Kisacanin and Brian Yu and Bryan Catanzaro and Carlo del Mundo and Chantal Hwang and Charles Wang and Cheng-Ping Hsieh and Chenghao Zhang and Chenhan Yu and Chetan Mungekar and Chintan Patel and Chris Alexiuk and Christopher Parisien and Collin Neale and Cyril Meurillon and Damon Mosk-Aoyama and Dan Su and Dane Corneil and Daniel Afrimi and Daniel Lo and Daniel Rohrer and Daniel Serebrenik and Daria Gitman and Daria Levy and Darko Stosic and David Mosallanezhad and Deepak Narayanan and Dhruv Nathawani and Dima Rekesh and Dina Yared and Divyanshu Kakwani and Dong Ahn and Duncan Riach and Dusan Stosic and Edgar Minasyan and Edward Lin and Eileen Long and Eileen Peters Long and Elad Segal and Elena Lantz and Ellie Evans and Elliott Ning and Eric Chung and Eric Harper and Eric Tramel and Erick Galinkin and Erik Pounds and Evan Briones and Evelina Bakhturina and Evgeny Tsykunov and Faisal Ladhak and Fay Wang and Fei Jia and Felipe Soares and Feng Chen and Ferenc Galko and Frank Sun and Frankie Siino and Gal Hubara Agam and Ganesh Ajjanagadde and Gantavya Bhatt and Gargi Prasad and George Armstrong and Gerald Shen and Gorkem Batmaz and Grigor Nalbandyan and Haifeng Qian and Harsh Sharma and Hayley Ross and Helen Ngo and Herbert Hum and Herman Sahota and Hexin Wang and Himanshu Soni and Hiren Upadhyay and Huizi Mao and Huy C Nguyen and Huy Q Nguyen and Iain Cunningham and Ido Galil and Ido Shahaf and Igor Gitman and Ilya Loshchilov and Itamar Schen and Itay Levy and Ivan Moshkov and Izik Golan and Izzy Putterman and Jan Kautz and Jane Polak Scowcroft and Jared Casper and Jatin Mitra and Jeffrey Glick and Jenny Chen and Jesse Oliver and Jian Zhang and Jiaqi Zeng and Jie Lou and Jimmy Zhang and Jinhang Choi and Jining Huang and Joey Conway and Joey Guman and John Kamalu and Johnny Greco and Jonathan Cohen and Joseph Jennings and Joyjit Daw and Julien Veron Vialard and Junkeun Yi and Jupinder Parmar and Kai Xu and Kan Zhu and Kari Briski and Katherine Cheung and Katherine Luna and Keith Wyss and Keshav Santhanam and Kevin Shih and Kezhi Kong and Khushi Bhardwaj and Kirthi Shankar and Krishna C. Puvvada and Krzysztof Pawelec and Kumar Anik and Lawrence McAfee and Laya Sleiman and Leon Derczynski and Li Ding and Lizzie Wei and Lucas Liebenwein and Luis Vega and Maanu Grover and Maarten Van Segbroeck and Maer Rodrigues de Melo and Mahdi Nazemi and Makesh Narsimhan Sreedhar and Manoj Kilaru and Maor Ashkenazi and Marc Romeijn and Marcin Chochowski and Mark Cai and Markus Kliegl and Maryam Moosaei and Matt Kulka and Matvei Novikov and Mehrzad Samadi and Melissa Corpuz and Mengru Wang and Meredith Price and Michael Andersch and Michael Boone and Michael Evans and Miguel Martinez and Mikail Khona and Mike Chrzanowski and Minseok Lee and Mohammad Dabbah and Mohammad Shoeybi and Mostofa Patwary and Nabin Mulepati and Najeeb Nabwani and Natalie Hereth and Nave Assaf and Negar Habibi and Neta Zmora and Netanel Haber and Nicola Sessions and Nidhi Bhatia and Nikhil Jukar and Nikki Pope and Nikolai Ludwig and Nima Tajbakhsh and Nir Ailon and Nirmal Juluru and Nishant Sharma and Oleksii Hrinchuk and Oleksii Kuchaiev and Olivier Delalleau and Oluwatobi Olabiyi and Omer Ullman Argov and Omri Puny and Oren Tropp and Ouye Xie and Parth Chadha and Pasha Shamis and Paul Gibbons and Pavlo Molchanov and Pawel Morkisz and Peter Dykas and Peter Jin and Pinky Xu and Piotr Januszewski and Pranav Prashant Thombre and Prasoon Varshney and Pritam Gundecha and Przemek Tredak and Qing Miao and Qiyu Wan and Rabeeh Karimi Mahabadi and Rachit Garg and Ran El-Yaniv and Ran Zilberstein and Rasoul Shafipour and Rich Harang and Rick Izzo and Rima Shahbazyan and Rishabh Garg and Ritika Borkar and Ritu Gala and Riyad Islam and Robert Hesse and Roger Waleffe and Rohit Watve and Roi Koren and Ruoxi Zhang and Russell Hewett and Russell J. Hewett and Ryan Prenger and Ryan Timbrook and Sadegh Mahdavi and Sahil Modi and Samuel Kriman and Sangkug Lim and Sanjay Kariyappa and Sanjeev Satheesh and Saori Kaji and Satish Pasumarthi and Saurav Muralidharan and Sean Narentharen and Sean Narenthiran and Seonmyeong Bak and Sergey Kashirsky and Seth Poulos and Shahar Mor and Shanmugam Ramasamy and Shantanu Acharya and Shaona Ghosh and Sharath Turuvekere Sreenivas and Shelby Thomas and Shiqing Fan and Shreya Gopal and Shrimai Prabhumoye and Shubham Pachori and Shubham Toshniwal and Shuoyang Ding and Siddharth Singh and Simeng Sun and Smita Ithape and Somshubra Majumdar and Soumye Singhal and Stas Sergienko and Stefania Alborghetti and Stephen Ge and Sugam Dipak Devare and Sumeet Kumar Barua and Suseella Panguluri and Suyog Gupta and Sweta Priyadarshi and Syeda Nahida Akter and Tan Bui and Teodor-Dumitru Ene and Terry Kong and Thanh Do and Tijmen Blankevoort and Tim Moon and Tom Balough and Tomer Asida and Tomer Bar Natan and Tomer Ronen and Tugrul Konuk and Twinkle Vashishth and Udi Karpas and Ushnish De and Vahid Noorozi and Vahid Noroozi and Venkat Srinivasan and Venmugil Elango and Victor Cui and Vijay Korthikanti and Vinay Rao and Vitaly Kurin and Vitaly Lavrukhin and Vladimir Anisimov and Wanli Jiang and Wasi Uddin Ahmad and Wei Du and Wei Ping and Wenfei Zhou and Will Jennings and William Zhang and Wojciech Prazuch and Xiaowei Ren and Yashaswi Karnati and Yejin Choi and Yev Meyer and Yi-Fu Wu and Yian Zhang and Yigong Qin and Ying Lin and Yonatan Geifman and Yonggan Fu and Yoshi Subara and Yoshi Suhara and Yubo Gao and Zach Moshe and Zhen Dong and Zhongbo Zhu and Zihan Liu and Zijia Chen and Zijie Yan},
      year={2025},
      eprint={2512.20856},
      archivePrefix={arXiv},
      primaryClass={cs.CL},
      url={https://arxiv.org/abs/2512.20856}, 
}

@misc{deepseekai2026deepseekv4,
      title={DeepSeek-V4: Towards Highly Efficient Million-Token Context Intelligence},
      author={DeepSeek-AI},
      year={2026},
}

@misc{kimiteam2026attentionresiduals,
      title={Attention Residuals}, 
      author={Kimi Team and Guangyu Chen and Yu Zhang and Jianlin Su and Weixin Xu and Siyuan Pan and Yaoyu Wang and Yucheng Wang and Guanduo Chen and Bohong Yin and Yutian Chen and Junjie Yan and Ming Wei and Y. Zhang and Fanqing Meng and Chao Hong and Xiaotong Xie and Shaowei Liu and Enzhe Lu and Yunpeng Tai and Yanru Chen and Xin Men and Haiqing Guo and Y. Charles and Haoyu Lu and Lin Sui and Jinguo Zhu and Zaida Zhou and Weiran He and Weixiao Huang and Xinran Xu and Yuzhi Wang and Guokun Lai and Yulun Du and Yuxin Wu and Zhilin Yang and Xinyu Zhou},
      year={2026},
      eprint={2603.15031},
      archivePrefix={arXiv},
      primaryClass={cs.CL},
      url={https://arxiv.org/abs/2603.15031}, 
}

@misc{choi2026kexaonetechnicalreport,
      title={K-EXAONE Technical Report}, 
      author={Eunbi Choi and Kibong Choi and Seokhee Hong and Junwon Hwang and Hyojin Jeon and Hyunjik Jo and Joonkee Kim and Seonghwan Kim and Soyeon Kim and Sunkyoung Kim and Yireun Kim and Yongil Kim and Haeju Lee and Jinsik Lee and Kyungmin Lee and Sangha Park and Heuiyeen Yeen and Hwan Chang and Stanley Jungkyu Choi and Yejin Choi and Jiwon Ham and Kijeong Jeon and Geunyeong Jeong and Gerrard Jeongwon Jo and Yonghwan Jo and Jiyeon Jung and Naeun Kang and Dohoon Kim and Euisoon Kim and Hayeon Kim and Hyosang Kim and Hyunseo Kim and Jieun Kim and Minu Kim and Myoungshin Kim and Unsol Kim and Youchul Kim and YoungJin Kim and Chaeeun Lee and Chaeyoon Lee and Changhun Lee and Dahm Lee and Edward Hwayoung Lee and Honglak Lee and Jinsang Lee and Jiyoung Lee and Sangeun Lee and Seungwon Lim and Solji Lim and Woohyung Lim and Chanwoo Moon and Jaewoo Park and Jinho Park and Yongmin Park and Hyerin Seo and Wooseok Seo and Yongwoo Song and Sejong Yang and Sihoon Yang and Chang En Yea and Sihyuk Yi and Chansik Yoon and Dongkeun Yoon and Sangyeon Yoon and Hyeongu Yun},
      year={2026},
      eprint={2601.01739},
      archivePrefix={arXiv},
      primaryClass={cs.CL},
      url={https://arxiv.org/abs/2601.01739}, 
}

@misc{park2026solaropentechnicalreport,
      title={Solar Open Technical Report},
      author={Sungrae Park and Sanghoon Kim and Jungho Cho and Gyoungjin Gim and Dawoon Jung and Mikyoung Cha and Eunhae Choo and Taekgyu Hong and Minbyul Jeong and SeHwan Joo and Minsoo Khang and Eunwon Kim and Minjeong Kim and Sujeong Kim and Yunsu Kim and Hyeonju Lee and Seunghyun Lee and Sukyung Lee and Siyoung Park and Gyungin Shin and Inseo Song and Wonho Song and Seonghoon Yang and Seungyoun Yi and Sanghoon Yoon and Jeonghyun Ko and Seyoung Song and Keunwoo Choi and Hwalsuk Lee and Sunghun Kim and Du-Seong Chang and Kyunghyun Cho and Junsuk Choe and Hwaran Lee and Jae-Gil Lee and KyungTae Lim and Alice Oh},
      year={2026},
      eprint={2601.07022},
      archivePrefix={arXiv},
      primaryClass={cs.CL},
      url={https://arxiv.org/abs/2601.07022}, 
}

@misc{tie2025surveyposttraininglargelanguage,
      title={A Survey on Post-training of Large Language Models}, 
      author={Guiyao Tie and Zeli Zhao and Dingjie Song and Fuyang Wei and Rong Zhou and Yurou Dai and Wen Yin and Zhejian Yang and Jiangyue Yan and Yao Su and Zhenhan Dai and Yifeng Xie and Yihan Cao and Lichao Sun and Pan Zhou and Lifang He and Hechang Chen and Yu Zhang and Qingsong Wen and Tianming Liu and Neil Zhenqiang Gong and Jiliang Tang and Caiming Xiong and Heng Ji and Philip S. Yu and Jianfeng Gao},
      year={2025},
      eprint={2503.06072},
      archivePrefix={arXiv},
      primaryClass={cs.CL},
      url={https://arxiv.org/abs/2503.06072}, 
}

@inproceedings{NIPS2000_44968aec,
 author = {Dugas, Charles and Bengio, Yoshua and B\'{e}lisle, Fran\c{c}ois and Nadeau, Claude and Garcia, Ren\'{e}},
 booktitle = {Advances in Neural Information Processing Systems},
 editor = {T. Leen and T. Dietterich and V. Tresp},
 pages = {},
 publisher = {MIT Press},
 title = {Incorporating Second-Order Functional Knowledge for Better Option Pricing},
 url = {https://proceedings.neurips.cc/paper_files/paper/2000/file/44968aece94f667e4095002d140b5896-Paper.pdf},
 volume = {13},
 year = {2000}
}

\appendix
\clearpage
\begin{figure}[htbp]
  \centering
  \begin{lstlisting}[language=python,basicstyle=\ttfamily\small]
h = Encoder(x)
sampled = RandomSampler(B, K)
for b in range(B):
  # Supervised step
  if b in sampled:
    h_prev = h
    h_prop = Reasoner(h)
    h = Gate(h_prev, h_prop)
    q_logit = ConfidenceHead(h)
    y_hat = Decoder(h)
    y_prev = Decoder(h_prev)
    loss = CE(y_hat, y_true)  # L_LM
    loss += SiLU(CE(y_hat, y_true)
                 - CE(y_prev, y_true))  # L_Mono
    loss += BCEWithLogits(q_logit,
                          (y_hat == y_true))  # L_Q
    loss.backward()
    opt.step()
    opt.zero_grad()
    h = h.detach()
  else:  # Unsupervised step
    with no_grad():
      h = Gate(h, Reasoner(h))
    h = h.detach()
  \end{lstlisting}
  \vspace{0.5\baselineskip}
  \caption{Pseudocode of LoopUS.}
  \label{fig:pseudocode}
\end{figure}

\section{Experimental Details}
\label{appendix:experimental_details}

\subsection{Backbones and Training Data}
We evaluate Qwen3-1.7B, Qwen3-4B, Qwen3-8B, TinyLlama, and Phi-4 backbones~\cite{yang2025qwen3technicalreport,abdin2024phi4technicalreport}. Our reported main experiments use streaming training on FineWeb-Edu with the CC-MAIN-2025-26 configuration~\citep{penedo2024the}, a total budget of 3B tokens, and sequence length 1024. The released public reference recipes are built on the same data pipeline. When applying the LoopUS architecture, the models are unrolled by selecting specific layers as the encoder and decoder, reserving the intermediate layers as the reusable reasoning block:
\begin{itemize}
    \item \textbf{Qwen3-1.7B:} Encoder layers 0--1, Decoder layer 27.
    \item \textbf{Qwen3-4B:} Encoder layers 0--1, Decoder layer 35.
    \item \textbf{Qwen3-8B:} Encoder layers 0--5, Decoder layer 35.
    \item \textbf{Phi-4:} Encoder layers 0--5, Decoder layer 39.
    \item \textbf{TinyLlama:} Encoder layer 0, Decoder layer 21.
\end{itemize}
This formulation separates a shallow encoder from a late decoder while repurposing the entire middle transformer block as the looped latent workspace.

\subsection{Optimization Details}

Training is implemented with Accelerate and distributed data parallelism, with gradient checkpointing enabled throughout the run. The reference script uses AdamW with a learning rate of $5\times10^{-5}$, one epoch over the token budget, bf16 mixed precision, FlashAttention-2~\cite{dao2024flashattention}, cosine scheduling, 300 warmup steps, 8 dataloader workers, and pinned-memory data loading. Logging is performed every 50 steps, checkpoints are saved every 5000 steps, and at most 3 checkpoints are retained. The data pipeline reserves a small held-out slice with $\texttt{val\_ratio}=10^{-4}$, although periodic validation is disabled in the released main training script by setting $\texttt{eval\_interval}=-1$.

The loop-specific configuration uses 20 total reasoning steps with deep supervision on 5 loop positions per example, a training stopping threshold of 0.55, and the \texttt{all} stopping mode. The training code also supports checkpoint-time lm-evaluation-harness runs; in the released script this option is enabled for WikiText with a limit of 200 samples.

\subsection{Evaluation Details}
We evaluate five pretrained backbones: Qwen3-1.7B, Qwen3-4B, Qwen3-8B, TinyLlama, and Phi-4. The reported training runs were conducted in cloud GPU environments, using NVIDIA L40S GPUs for Qwen3-1.7B and TinyLlama, NVIDIA RTX PRO 6000 GPUs for Qwen3-4B and Qwen3-8B, and NVIDIA H200 GPUs for Phi-4. Unless otherwise stated, models are trained on FineWeb-Edu with a budget of 3B tokens, context length 1024, AdamW, cosine learning-rate decay, bf16 mixed precision, and the default LoopUS setting of $B{=}20$ total loop steps with $K{=}5$ supervised depths per batch.

For evaluation, we use \texttt{lm-evaluation-harness} in a zero-shot setting and report perplexity on WikiText and Lambada, together with accuracy on MMLU, HellaSwag, ARC-Easy, ARC-Challenge, PIQA, WinoGrande, and OpenBookQA. Standard inference uses a maximum recursion budget of 8, a stopping threshold of 0.6, and KV caching for autoregressive decoding. The main experiments are organized to test three claims: that architectural recasting improves the pretrained backbone while preserving its language-modeling interface, that selective gating and random deep supervision improve adaptation behavior under limited budgets relative to prior retrofitting schemes, and that the resulting recurrence behaves as controlled iterative refinement rather than uncontrolled extra depth.

\subsection{KV-Cache Implementation for Autoregressive Inference}

\begin{figure}[ht!]
\centering
\includegraphics[width=\textwidth]{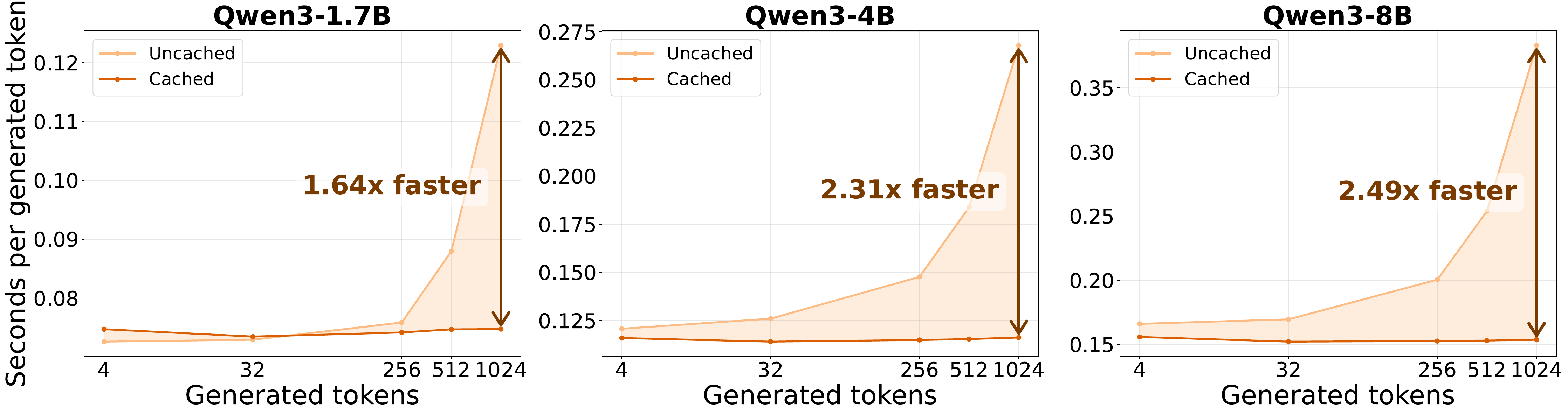}
\caption{Effect of KV caching on LoopUS autoregressive decoding speed with the recursion budget fixed to $B=8$. Across Qwen3-1.7B, Qwen3-4B, and Qwen3-8B, caching consistently reduces seconds per generated token, with the largest gains appearing at longer generations.}
\label{fig:cached_speed_comparison}
\end{figure}

Let $x_{1:t}$ denote the already processed prefix at autoregressive decoding step $t$, and let $B$ denote the maximum recursion budget used for that forward pass. We use $t$ for decoding time and $b$ for latent refinement depth. LoopUS stores the autoregressive cache state as
\begin{equation}
\mathcal{C}_t = \left(\mathcal{C}^{\mathrm{enc}}_t, \{\mathcal{C}^{\mathrm{rea},b}_t\}_{b=1}^{B}, \mathcal{C}^{\mathrm{dec}}_t, s_t\right),
\end{equation}
where $\mathcal{C}^{\mathrm{enc}}_t$ is the encoder KV cache, $\mathcal{C}^{\mathrm{rea},b}_t$ is the KV cache of the $b$-th refinement depth, $\mathcal{C}^{\mathrm{dec}}_t$ is the decoder KV cache, and $s_t$ is the number of previously seen tokens. In the implementation, these objects are instantiated as separate Hugging Face \texttt{DynamicCache}s plus a \texttt{seen\_tokens} counter. Conceptually, inference proceeds in two phases. In the \emph{prefill} phase, the full prompt is processed once and all encoder, reasoning, and decoder caches are populated for the prompt tokens. In the subsequent \emph{decode} phase, LoopUS no longer recomputes the full prefix. Instead, when generating token $x_{t+1}$, each module receives only the newest token representation together with its previously accumulated cache, so the new query attends to the stored keys and values of the entire prefix while appending only one additional KV entry per layer and per loop depth. The initial forward pass processes the full prompt, whereas every later decoding step uses only the newest token,
\begin{equation}
x_{t+1}, \qquad p_{t+1} = s_t,
\end{equation}
with the absolute position $p_{t+1}$ supplied through \texttt{cache\_position}. This keeps rotary position IDs and causal masks aligned with the full prefix even though the model no longer recomputes $x_{1:t}$.

The per-token latent update can be written as
\begin{align}
h^{(0)}_{t+1}, \mathcal{C}^{\mathrm{enc}}_{t+1}
&= \mathcal{E}\left(x_{t+1}; \mathcal{C}^{\mathrm{enc}}_t, p_{t+1}\right), \\
h^{(b)}_{t+1}, \mathcal{C}^{\mathrm{rea},b}_{t+1}
&= \mathcal{R}_b\left(h^{(b-1)}_{t+1}; \mathcal{C}^{\mathrm{rea},b}_t, p_{t+1}\right), \quad b = 1, \dots, B, \\
\ell_{t+1}, \mathcal{C}^{\mathrm{dec}}_{t+1}
&= \mathcal{D}\left(h^{(B)}_{t+1}; \mathcal{C}^{\mathrm{dec}}_t, p_{t+1}\right),
\end{align}
where $\ell_{t+1}$ denotes the next-token logits. The key design choice is that LoopUS does not share a single cache across all recursive refinements; instead, each refinement depth $b$ has its own cache $\mathcal{C}^{\mathrm{rea},b}_t$. This separation is necessary because the hidden state entering loop depth $b$ is not the same object as the hidden state entering depth $b' \neq b$: although the block parameters are shared, the token representations evolve after every refinement step, so the corresponding keys and values represent different latent trajectories. Reusing one common cache across all depths would mix KV states produced at different refinement stages and would no longer correspond to the actual recurrence being executed. Maintaining one cache per depth therefore preserves the semantics of the unrolled loop while still amortizing prefix computation over autoregressive time. This design matches the fact that the same middle block is reused as a depth-indexed dynamical operator rather than as one flat transformer pass, and it keeps the implementation compatible with recursive decoding views similar to Bae et al.~\citep{bae2025relaxedrecursive,bae2025mixtureofrecursions}.

If the active context reaches the maximum window size $M$, the cached state is truncated before appending the next token,
\begin{equation}
\mathcal{C}_t \leftarrow \operatorname{crop}(\mathcal{C}_t, M-1),
\qquad
s_t \leftarrow M-1,
\end{equation}
so all encoder, reasoning, and decoder caches remain synchronized with the same sliding window. In this form, KV caching turns LoopUS generation into an incremental state update over $\mathcal{C}_t$ rather than repeated recomputation of the entire prefix at every token.

To quantify the practical effect of this design, we benchmark autoregressive decoding with the recursion budget fixed to $B=8$ and measure token-generation throughput over five repeated runs on an NVIDIA L40S GPU. Figure~\ref{fig:cached_speed_comparison} summarizes the resulting cached-versus-uncached comparison across output lengths. At 1024 generated tokens, KV caching yields a $1.64\times$ speedup for Qwen3-1.7B, a $2.31\times$ speedup for Qwen3-4B, and a $2.49\times$ speedup for Qwen3-8B. This pattern is consistent with the observation of \citet{bae2025mixtureofrecursions} that looped transformers benefit strongly from cache reuse because recurrent decoding repeatedly reaccesses shared memory across refinement steps. Once the prefix state has been materialized, caching removes much of the repeated computation that would otherwise be spent reconstructing the same context at each autoregressive step.

\subsection{Halting and Recursion Diagnostics}
The halting comparison in Figure~\ref{fig:halting_comparison} is implemented with a fixed Qwen3-1.7B LoopUS checkpoint and compares three inference rules: threshold halting, convergence halting based on hidden-state change, and CDF-based halting. These runs use the task set \{MMLU, HellaSwag, ARC-Easy, PIQA, WinoGrande, Lambada, WikiText\}, maximum recursion budget $B=10$, batch size 16, maximum length 1024, seed 2026, and runtime-stat logging. The convergence sweep uses $\epsilon \in \{0.1, 1, 5, 10, 100\}$, while the CDF sweep uses thresholds $\{0.2, 0.3, 0.5, 0.7, 0.8\}$.

The separate recursion-depth study varies the inference budget on the Qwen3-4B LoopUS checkpoint while keeping the remaining evaluation settings fixed. This diagnostic isolates how far additional loop depth can improve performance before gains saturate or begin to trade off against over-refinement; the corresponding results are reported in Figure~\ref{fig:qwen4b_recursion_task_trends} in Section~\ref{sec:recursion_depth_diagnostics}.

\section{Dynamical and Geometric Interpretation of LoopUS}
\label{appendix:energy_based_interpretation}

Recent work suggests that latent-space iterative reasoning can be understood through a dynamical lens. \citet{gladstone2026energybased} argue that ``thinking'' can be framed as optimization with respect to a learned verifier that measures the compatibility between an input and a candidate prediction, with additional computation corresponding to iterative refinement until convergence. Complementarily, \citet{blondel2026autoregressive} establish an explicit bijection between autoregressive models and EBMs in function space. They show that a sequence-level energy decomposes into per-token rewards,
\begin{equation}
R(x, y) = \sum_{t=1}^{|y|} r(x \oplus y_{<t}, y_t),
\end{equation}
and that an autoregressive model's next-token logits $q$ relate to these per-token rewards $r$ through the soft Bellman equation: $q(s_t, y_t) = r(s_t, y_t) + V_q(s_t \oplus y_t)$, where $V_q$ is a soft value function encoding lookahead over all possible continuations. Thus even a model trained purely on next-token prediction implicitly induces a global compatibility function rather than making only isolated token-level decisions. Taken together, these results suggest that an autoregressive transformer can exhibit gradual refinement trajectories similar to energy minimization processes.

This view is consistent with LoopUS because our hidden-state analysis in Figure~\ref{fig:hidden_analysis_original_LLM} suggests that the reused middle layers already form a near-fixed-point latent workspace. LoopUS then turns this workspace into an iterative refinement process. Using the recursion-depth index $b$, after vectorizing the hidden state and writing $P^{(b)} = \mathrm{Diag}(\alpha^{(b)})$, Eq.~\ref{eq:loopus_gate} can be rewritten exactly as
\begin{equation}
h^{(b+1)} = h^{(b)} - P^{(b)}\left(h^{(b)} - \mathcal{M}(h^{(b)})\right).
\label{eq:appendix_relaxed_fp}
\end{equation}
This is a diagonally preconditioned relaxed fixed-point iteration. Therefore, the mathematically unconditional statement is that LoopUS searches for a latent state $h^{\star}$ satisfying $h^{\star} = \mathcal{M}(h^{\star})$. Any converged point with strictly positive diagonal entries in $P^{(b)}$ must satisfy this fixed-point condition. In other words, the rigorous part of the argument is a stable fixed-point search in latent space; the energy interpretation becomes exact only under an additional assumption on the residual field.

Specifically, suppose there exists a scalar potential $\Phi_x(h)$ on the relevant latent manifold such that
\begin{equation}
\nabla_h \Phi_x(h) = h - \mathcal{M}(h).
\label{eq:appendix_potential}
\end{equation}
Then Eq.~\ref{eq:appendix_relaxed_fp} becomes
\begin{equation}
h^{(b+1)} = h^{(b)} - P^{(b)} \nabla_h \Phi_x(h^{(b)}),
\end{equation}
which is simply diagonally preconditioned gradient descent on $\Phi_x$. If $\Phi_x$ is $L$-smooth and the effective step size is small enough, e.g., $\|P^{(b)}\|_2 \le 1/L$, the descent lemma yields
\begin{equation}
\Phi_x(h^{(b+1)}) \le \Phi_x(h^{(b)}) - \langle g^{(b)}, P^{(b)} g^{(b)} \rangle + \frac{L}{2}\|P^{(b)} g^{(b)}\|_2^2,
\quad g^{(b)} = \nabla_h \Phi_x(h^{(b)}),
\end{equation}
so the potential decreases whenever the gated step is sufficiently small. Under this assumption, fixed points of $\mathcal{M}$ coincide with stationary points of $\Phi_x$. This provides a geometric intuition for why LoopUS can be viewed as mimicking an energy minimization process, provided the residual field $h - \mathcal{M}(h)$ locally resembles a gradient field.

More directly, our training objective enforces a next-token-prediction surrogate energy descent. Define
\begin{equation}
E_x(h) = \mathrm{CE}\!\left(\mathcal{D}(h), x_{2:T}\right),
\end{equation}
which is exactly the next-token prediction loss for the gold continuation under the decoder. Then the monotonicity loss from Section~\ref{sec:method_training_objective} can be written as
\begin{equation}
\mathcal{L}_{\mathrm{mono}}^{(b)} = \operatorname{SiLU}\!\left(E_x(h^{(b+1)}) - E_x(h^{(b)})\right).
\end{equation}
Because SiLU is positive for positive arguments and produces only small negative values for negative arguments, minimizing $\mathcal{L}_{\mathrm{mono}}^{(b)}$ penalizes positive increments in $E_x$ and mildly rewards negative increments. Therefore, even without proving that $h - \mathcal{M}(h)$ is the gradient of a global scalar potential, LoopUS is explicitly trained to make this decoder-induced surrogate energy approximately non-increasing along the loop. The confidence head complements this by stopping once the predicted benefit of further refinement becomes small. In this sense, the halting trends in Figure~\ref{fig:halting_comparison} and the smooth contraction-like trajectories in Figures~\ref{fig:appendix_pca} and~\ref{fig:appendix_distance} are not just qualitative analogies: they are the empirical signature of a loop trained to behave like a descent process on a task-aligned surrogate energy.

The monotonicity loss and stepwise detachment sharpen this picture at the level of individual refinement transitions. The monotonicity term does not only prefer a good final state; it explicitly trains each local update $h^{(b)} \rightarrow h^{(b+1)}$ to avoid increasing the decoder surrogate energy $E_x$. Stepwise detachment then prevents gradients from coupling all loop positions into one long backpropagation-through-time graph, so each supervised depth is optimized primarily as a local correction applied to the current latent state rather than as one stage in a globally entangled trajectory. Combined with the selective gate, this biases LoopUS toward sequences of small, progressively improving latent edits instead of brittle one-shot remappings. In this limited dynamical sense, LoopUS resembles deterministic iterative refinement procedures such as DDIM~\citep{song2021denoising} and recent language diffusion models~\citep{nie2025large}: each step is trained to move the representation toward a more task-aligned region while preserving stable multi-step evolution.

The diffusion analogy should not be overstated. LoopUS does not inject explicit noise, does not learn a reverse diffusion process or noise schedule, and does not optimize a score-matching objective. Strictly speaking, LoopUS is also not an explicit EBM. We do not learn a standalone scalar energy over all input-output pairs, nor do we run gradient descent on that scalar in output space. A more precise statement is that LoopUS is an \emph{implicit latent optimizer}: the refinement process is amortized into the recurrent hidden-state dynamics of a pretrained autoregressive transformer, approximating a steepest-descent-like process through repeated latent updates. The connection to diffusion or EBMs is therefore geometric and dynamical rather than probabilistic. This distinction matters, but the main conceptual point remains: what makes the loop useful is not merely extra depth, but the emergence of a compatibility-seeking dynamical process that keeps allocating compute until the representation settles into a more predictive state.

\section{Additional Qualitative Example}
\label{appendix:qualitative_example}
\begin{figure}[htbp!]
\centering
\includegraphics[width=0.9\textwidth]{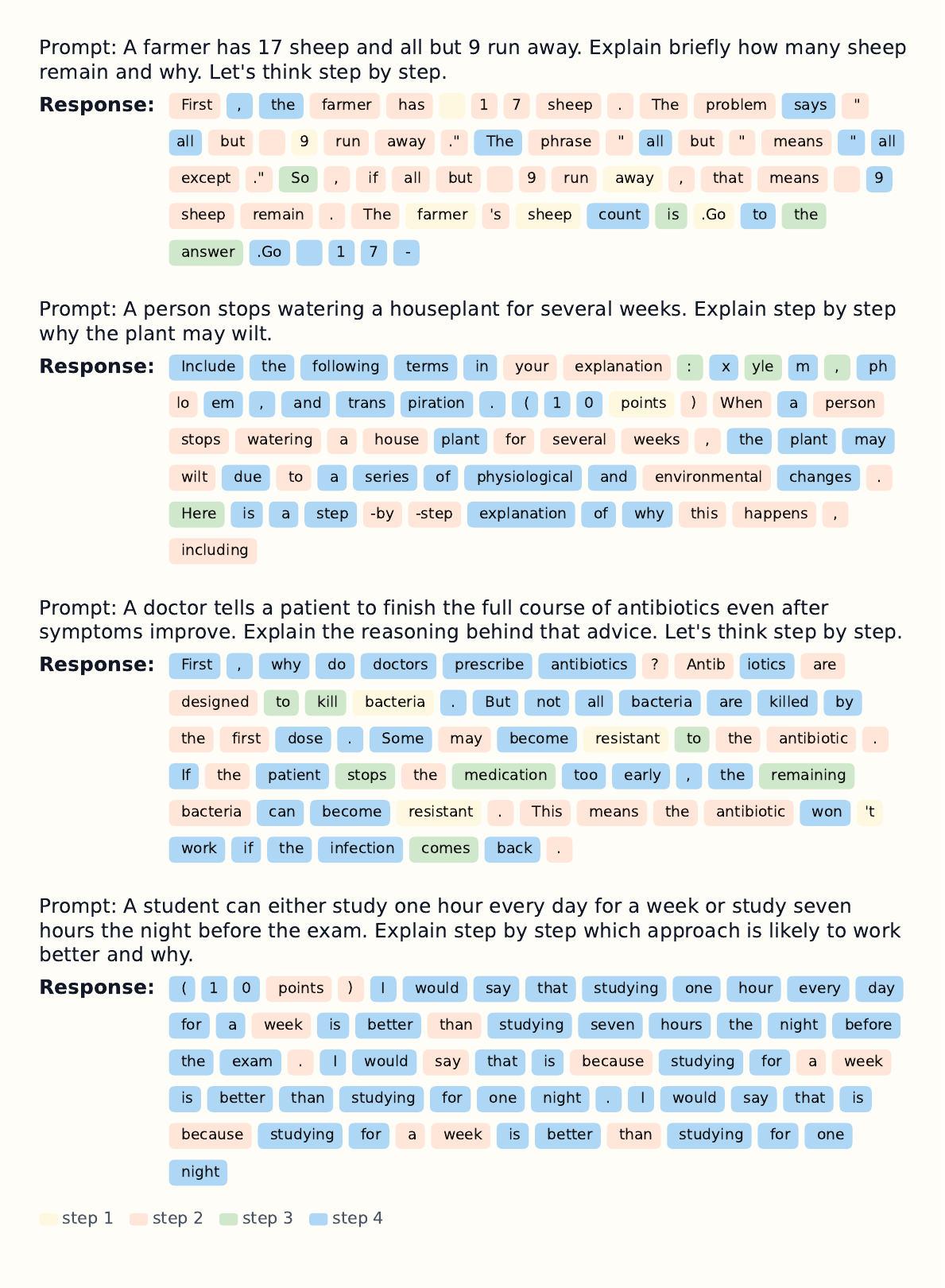}
\caption{Example generation thinking trace from LoopUS on Qwen3-4B. The figure visualizes how the model's intermediate reasoning trajectory evolves across loop iterations for a representative sample.}
\label{fig:appendix_thinking_trace}
\end{figure}

Figure~\ref{fig:appendix_thinking_trace} provides a representative example of how LoopUS refines its latent state across loop iterations during generation. Rather than making one large, brittle update, the model follows a coherent multi-step trajectory: early iterations make the largest representational corrections, while later iterations make smaller adjustments that stabilize the prediction. This qualitative behavior is consistent with the view developed in the main text that LoopUS acts as an iterative latent refinement process rather than as a one-shot remapping of the hidden state.

\begin{figure}[p]
\centering
\includegraphics[width=\textwidth,height=0.82\textheight,keepaspectratio]{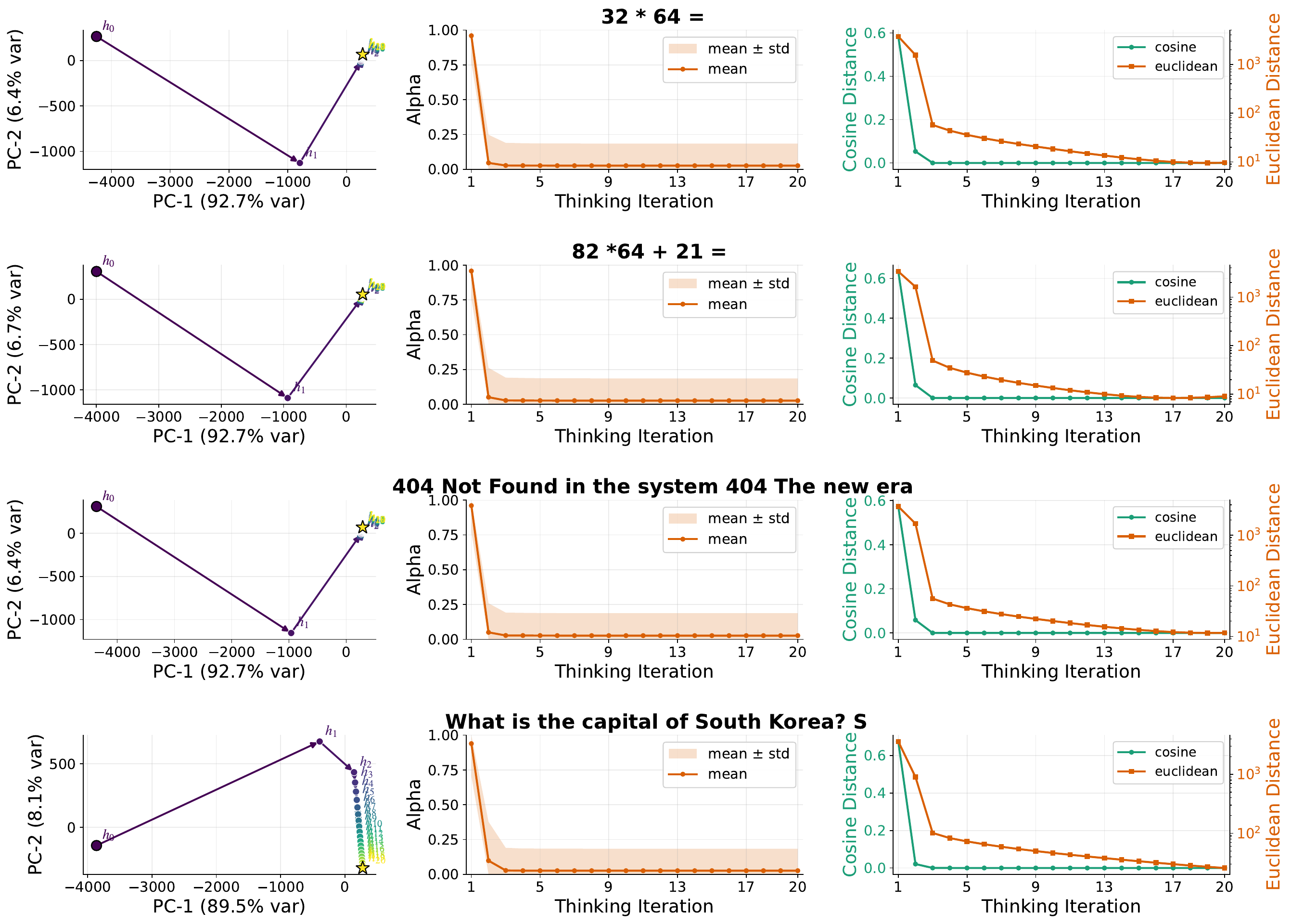}
\caption{LoopUS thinking PCA visualization for Qwen3-1.7B.}
\label{fig:appendix_loopus_thinking_pca_qwen3_1_7b}
\end{figure}

\begin{figure}[p]
\centering
\includegraphics[width=\textwidth,height=0.82\textheight,keepaspectratio]{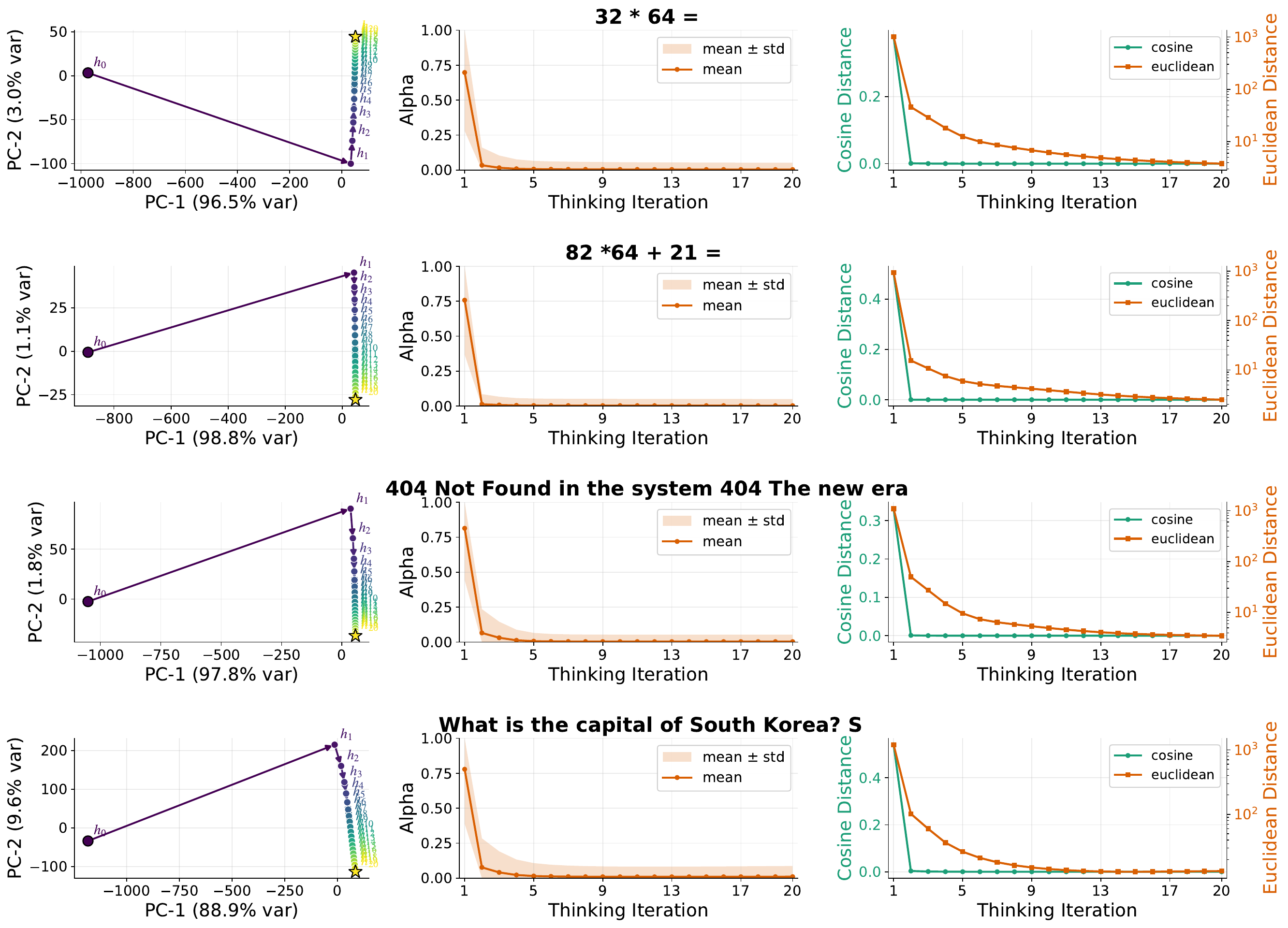}
\caption{LoopUS thinking PCA visualization for Qwen3-8B.}
\label{fig:appendix_loopus_thinking_pca_qwen3_8b}
\end{figure}

\begin{figure}[p]
\centering
\includegraphics[width=\textwidth,height=0.82\textheight,keepaspectratio]{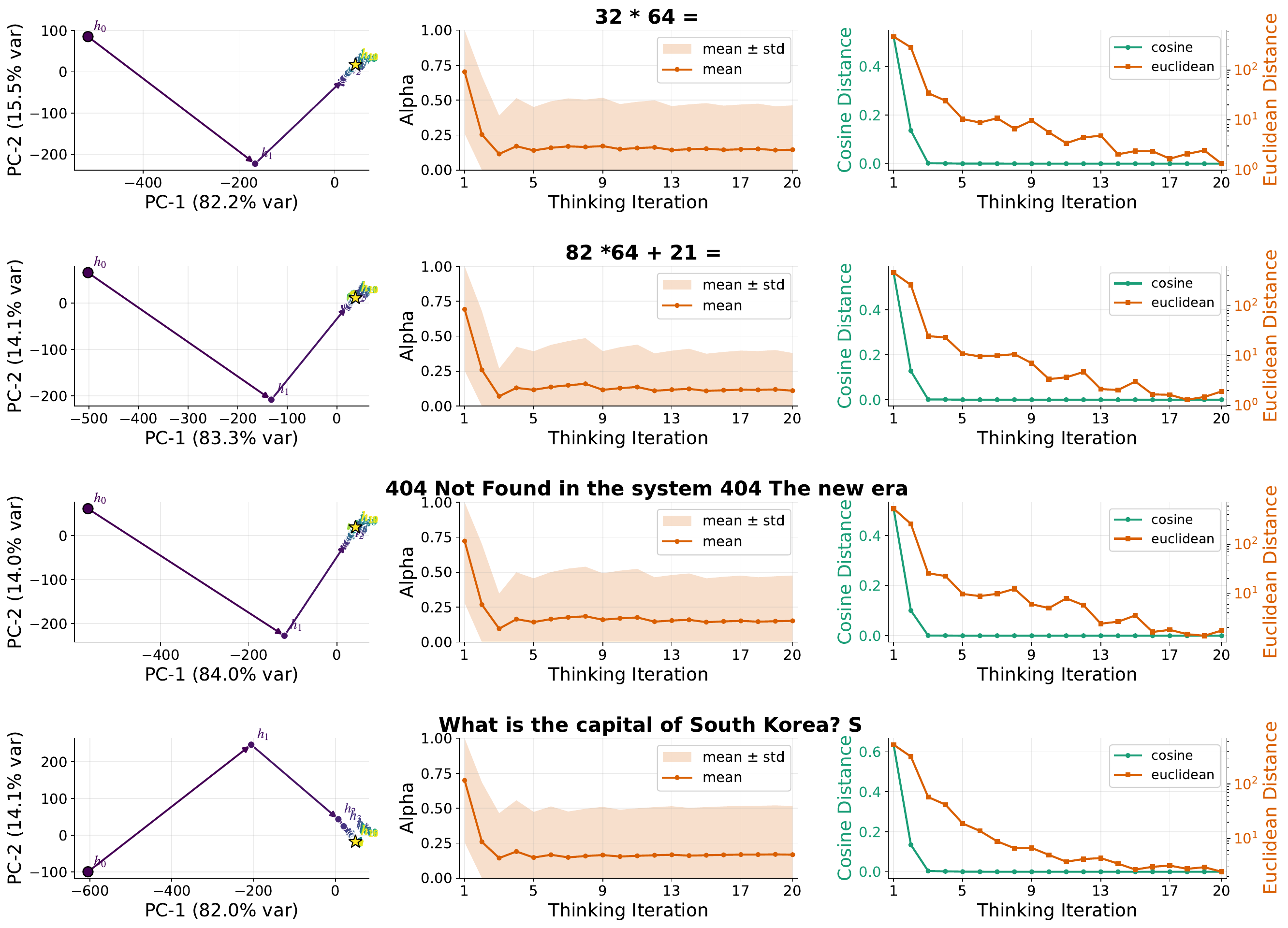}
\caption{LoopUS thinking PCA visualization for Qwen3-4B.}
\label{fig:appendix_loopus_thinking_pca_qwen4b}
\end{figure}

Figures~\ref{fig:appendix_loopus_thinking_pca_qwen3_1_7b}, \ref{fig:appendix_loopus_thinking_pca_qwen3_8b}, and \ref{fig:appendix_loopus_thinking_pca_qwen4b} further visualize these loop trajectories in a shared PCA space for multiple LoopUS backbones. Across models, the hidden states evolve along smooth, low-dimensional paths instead of exhibiting erratic jumps between iterations. The trajectories typically show larger motion in the first few refinement steps followed by progressive contraction, which supports our interpretation of LoopUS as a controlled descent-like process in latent space. Although the exact geometry varies by backbone, the common pattern is that iterative reuse of the reasoning block produces structured movement toward a more stable predictive state.

\section{Hidden-State Trajectory Analysis}
\label{appendix:hidden_state_trajectory_analysis}
\begin{figure}[ht!]
\centering
\begin{subfigure}[t]{0.49\textwidth}
	\centering
	\includegraphics[width=\textwidth]{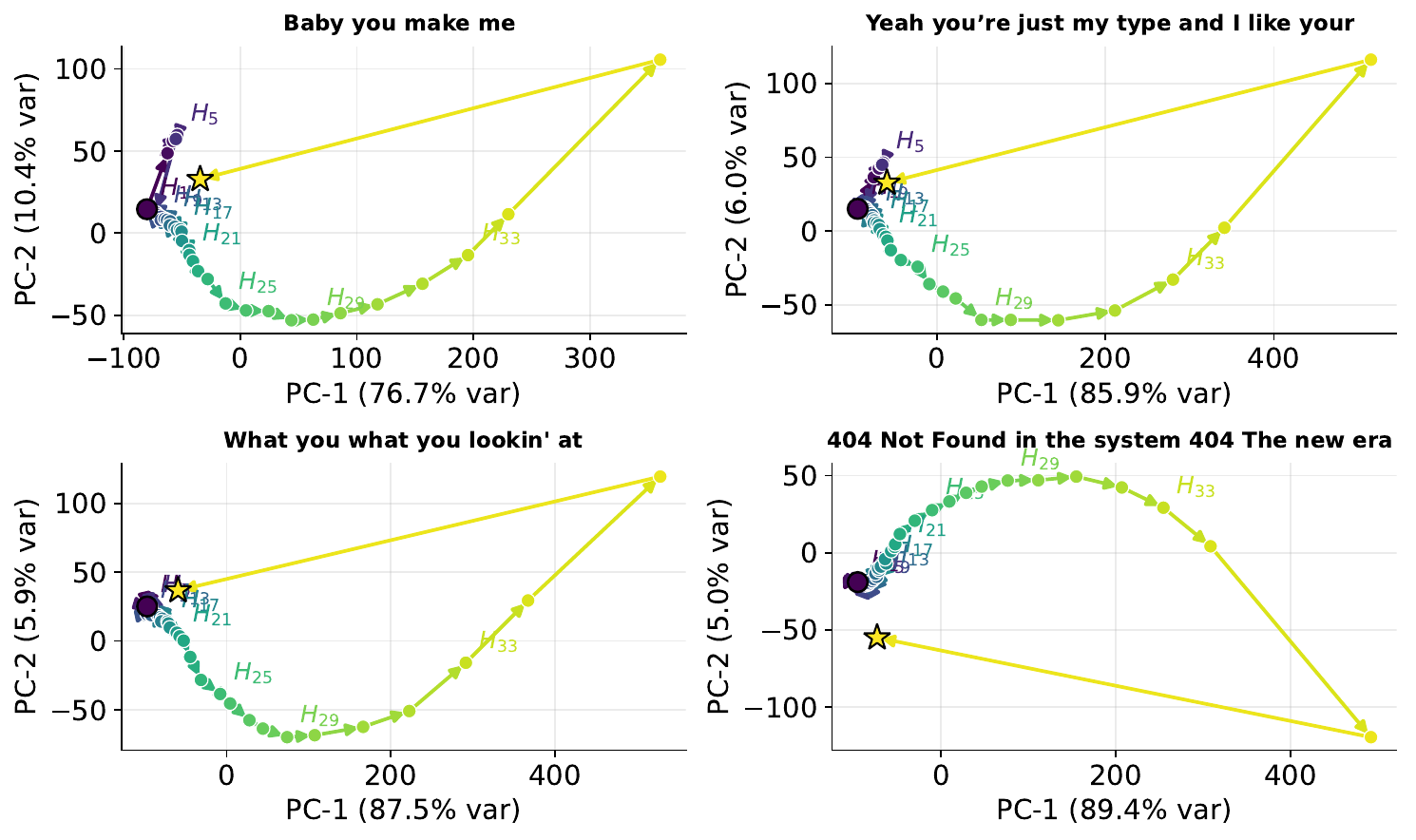}
	\caption{Qwen3-4B}
\end{subfigure}
\hfill
\begin{subfigure}[t]{0.49\textwidth}
	\centering
	\includegraphics[width=\textwidth]{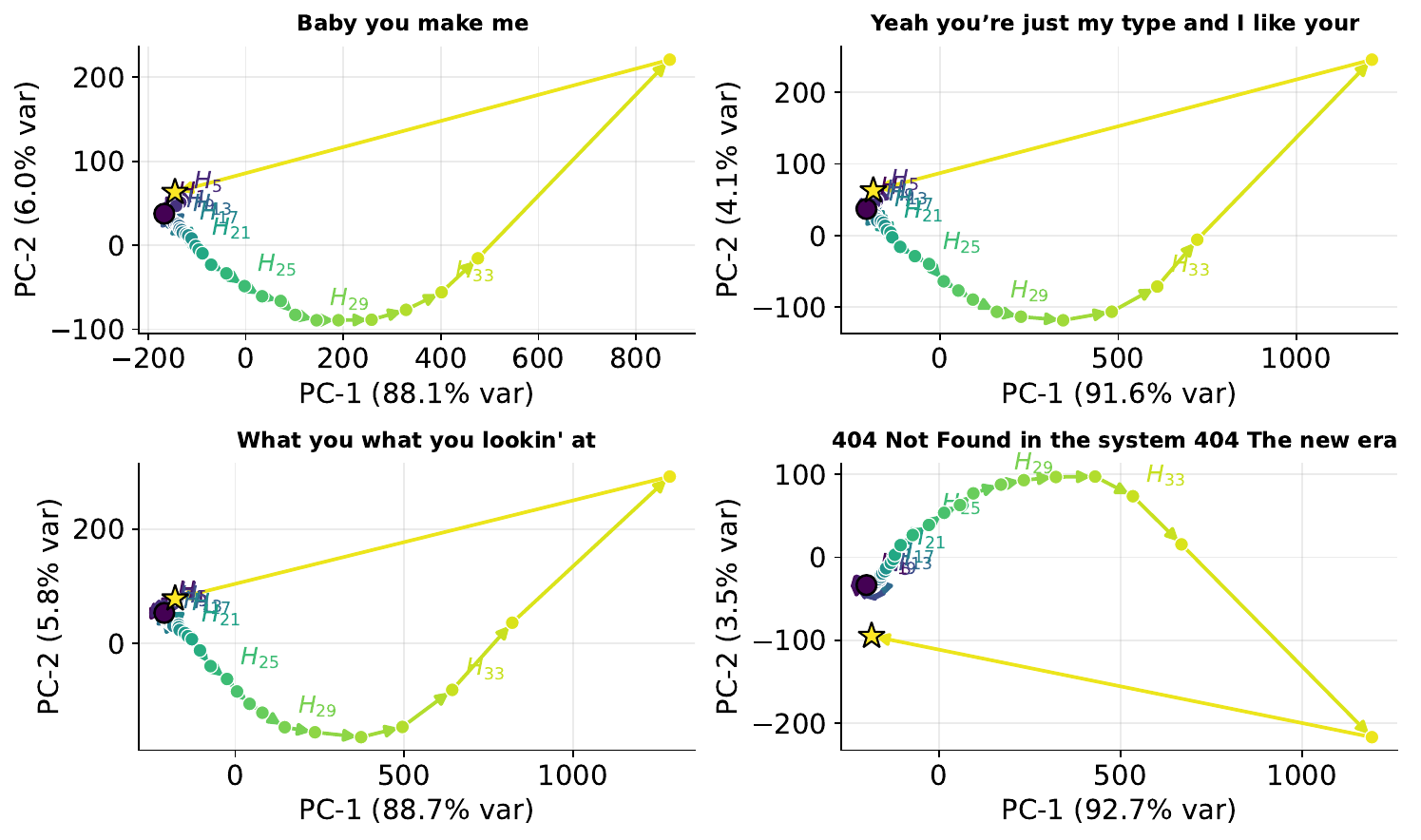}
	\caption{Qwen3-8B}
\end{subfigure}

\vspace{0.5em}

\begin{subfigure}[t]{0.49\textwidth}
	\centering
	\includegraphics[width=\textwidth]{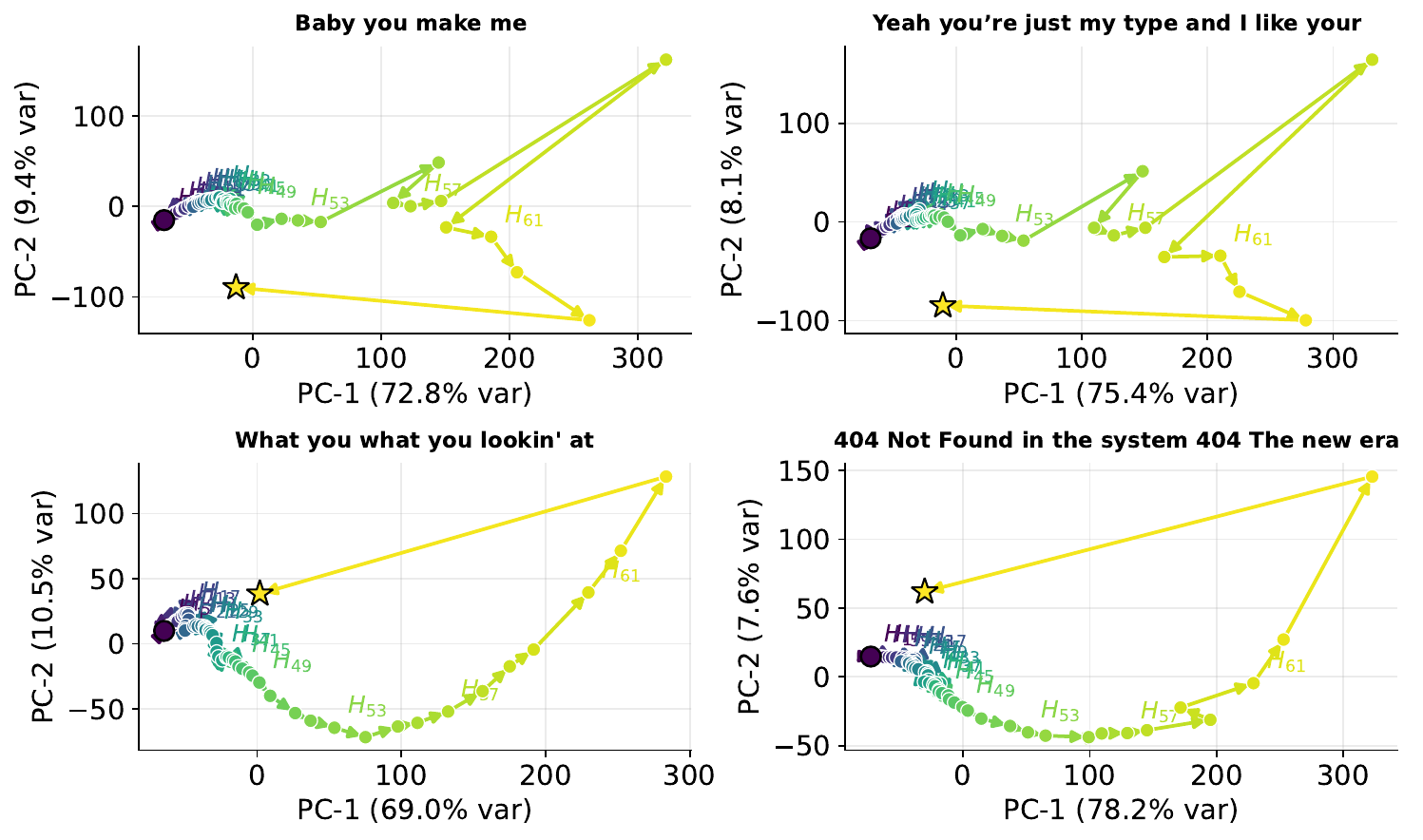}
	\caption{Qwen3.5-27B}
\end{subfigure}
\hfill
\begin{subfigure}[t]{0.49\textwidth}
	\centering
	\includegraphics[width=\textwidth]{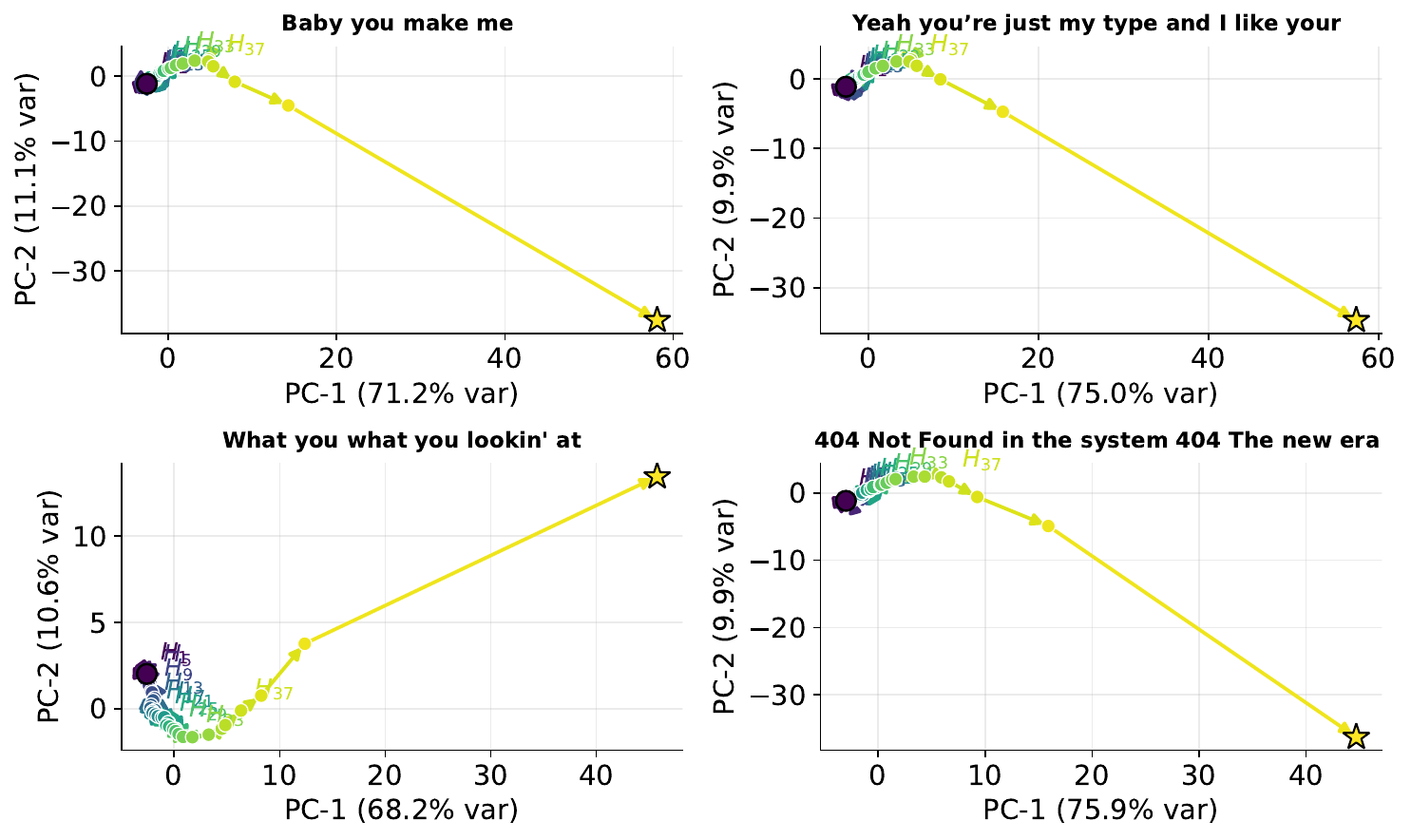}
	\caption{Qwen3.5-35B-A3B}
\end{subfigure}

\vspace{0.5em}

\begin{subfigure}[t]{0.49\textwidth}
	\centering
	\includegraphics[width=\textwidth]{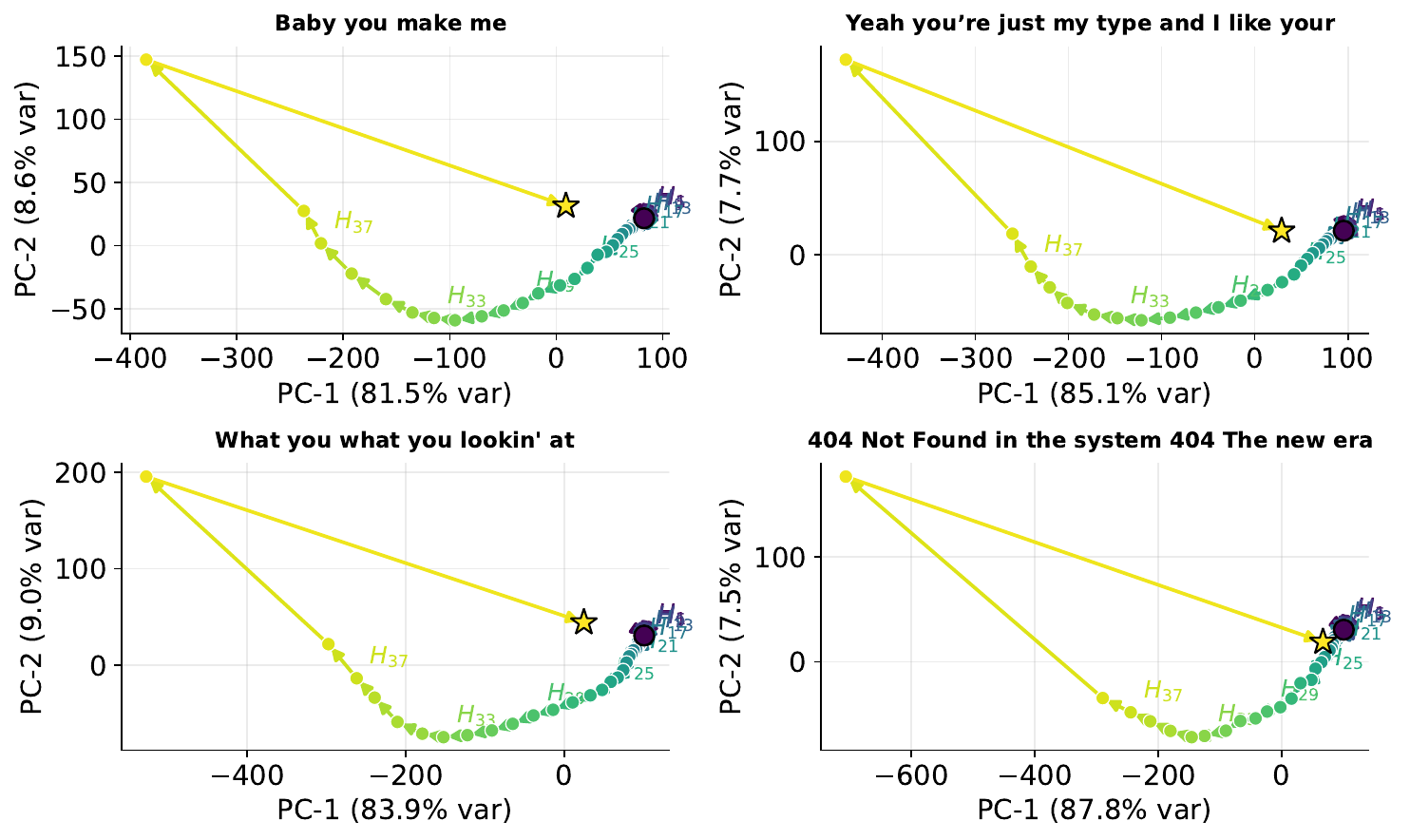}
	\caption{Phi-4}
\end{subfigure}
\hfill
\begin{subfigure}[t]{0.49\textwidth}
	\centering
	\includegraphics[width=\textwidth]{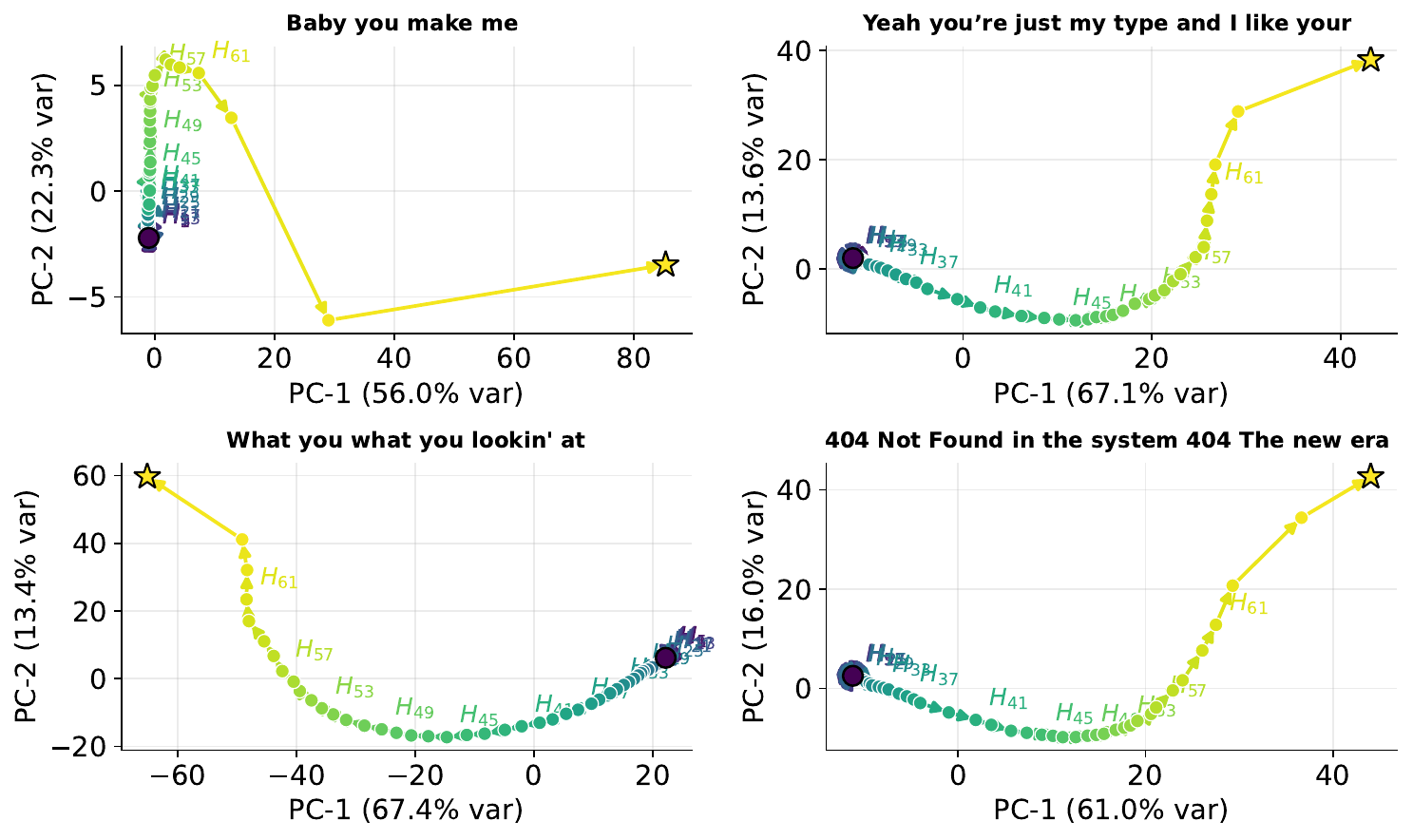}
	\caption{EXAONE-4.0-32B}
\end{subfigure}

\caption{PCA trajectories of hidden-state refinement across model backbones.}
\label{fig:appendix_pca}
\end{figure}

\begin{figure}[ht!]
\centering
\begin{subfigure}[t]{0.49\textwidth}
	\centering
	\includegraphics[width=\textwidth]{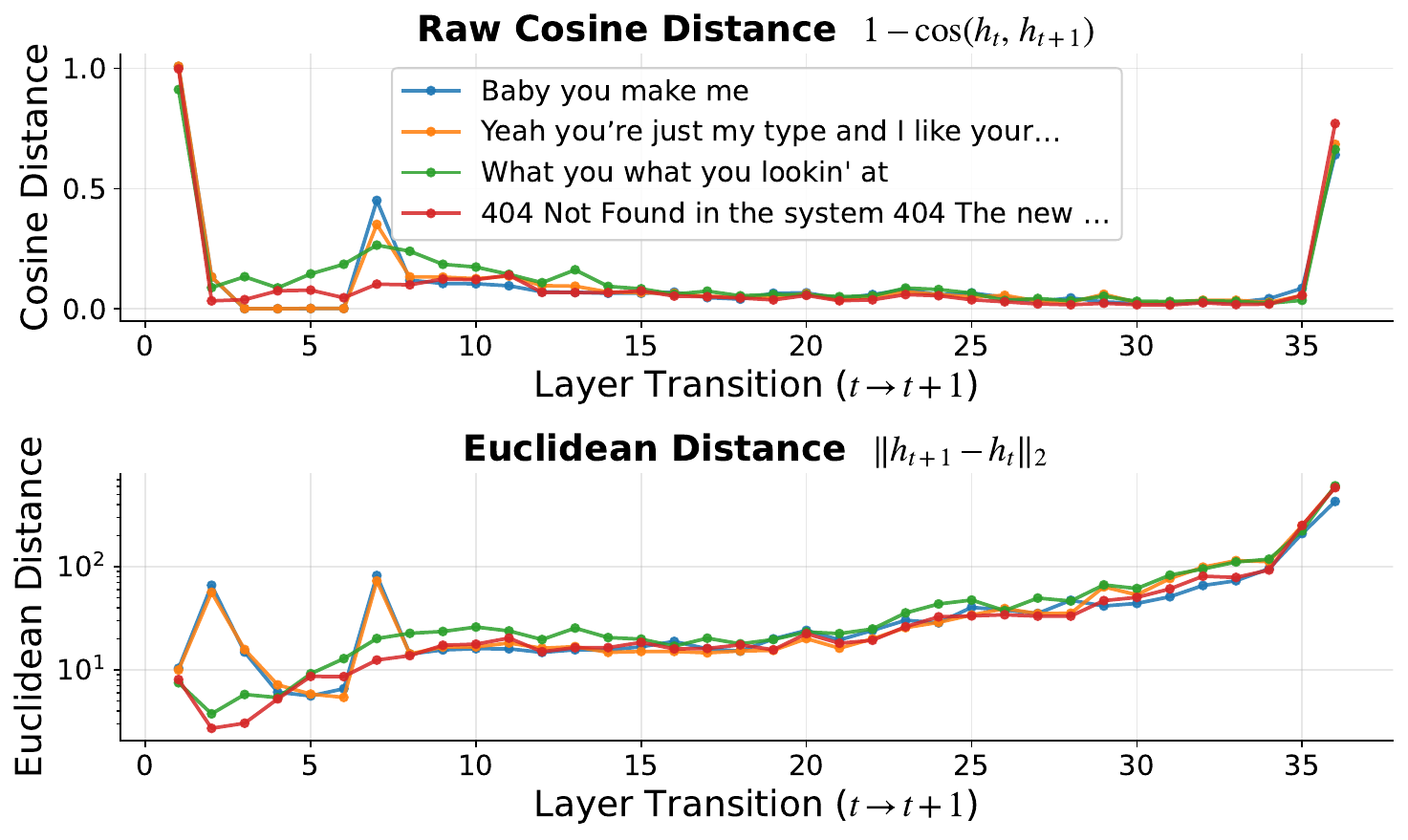}
	\caption{Qwen3-4B}
\end{subfigure}
\hfill
\begin{subfigure}[t]{0.49\textwidth}
	\centering
	\includegraphics[width=\textwidth]{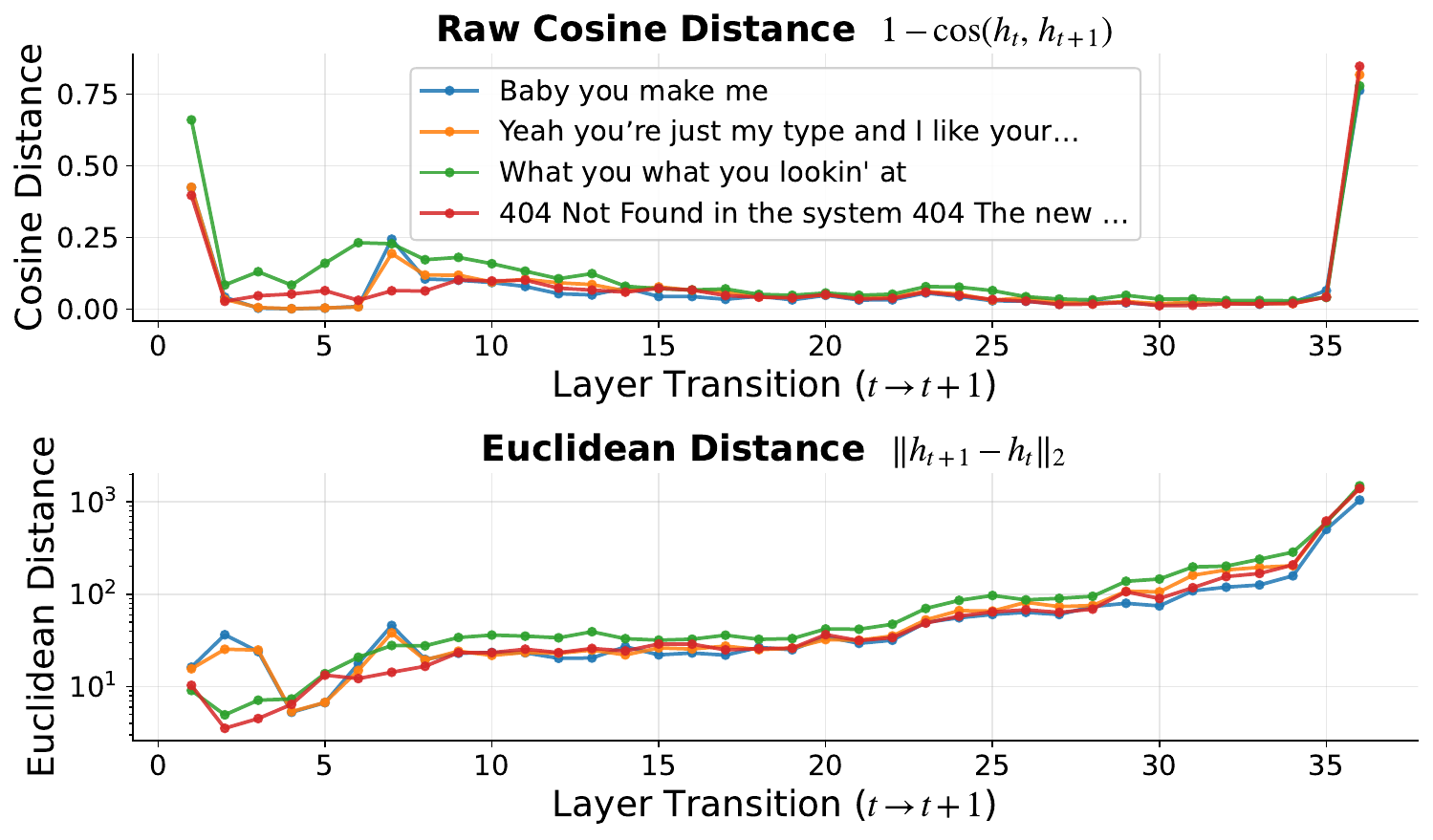}
	\caption{Qwen3-8B}
\end{subfigure}

\vspace{0.5em}

\begin{subfigure}[t]{0.49\textwidth}
	\centering
	\includegraphics[width=\textwidth]{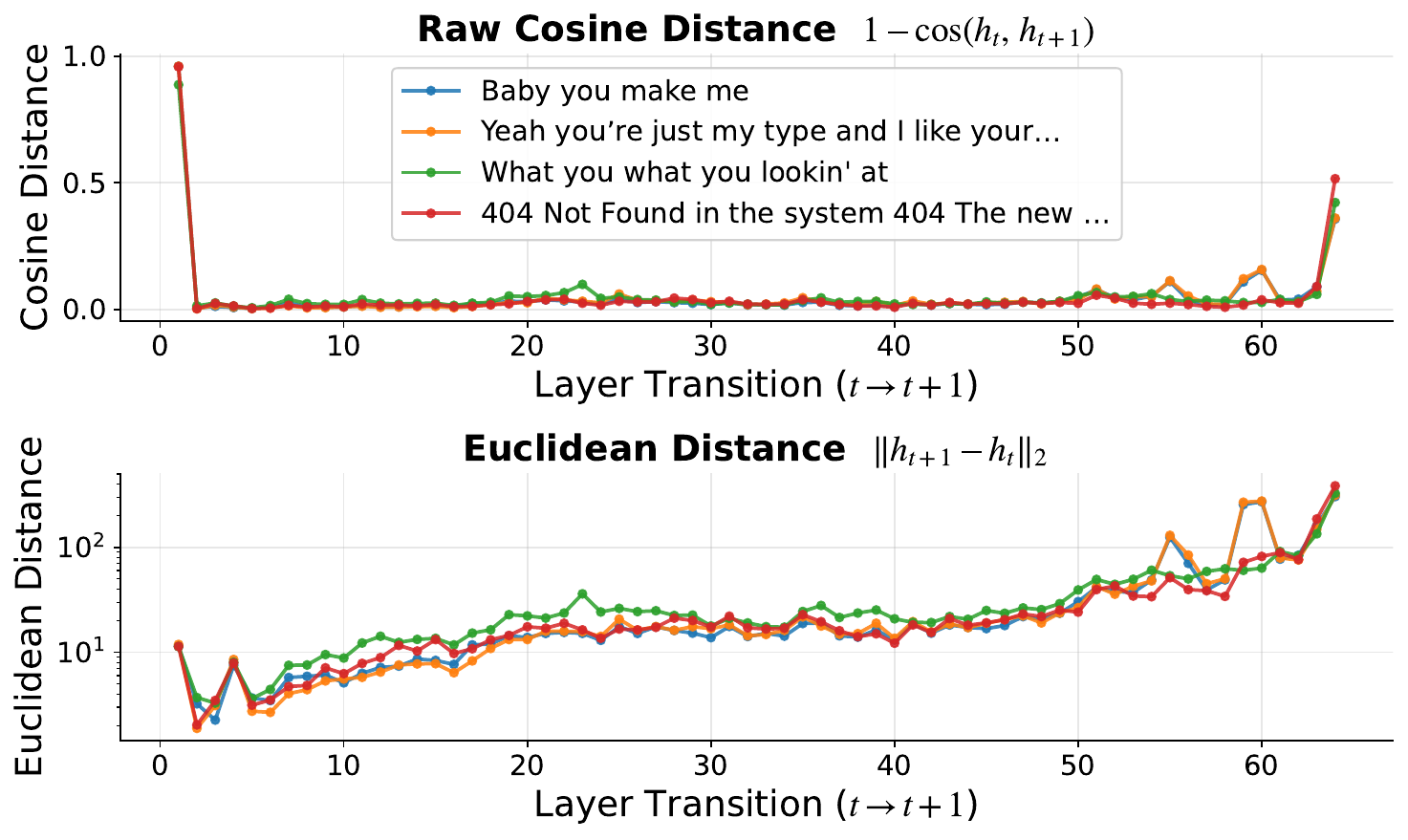}
	\caption{Qwen3.5-27B}
\end{subfigure}
\hfill
\begin{subfigure}[t]{0.49\textwidth}
	\centering
	\includegraphics[width=\textwidth]{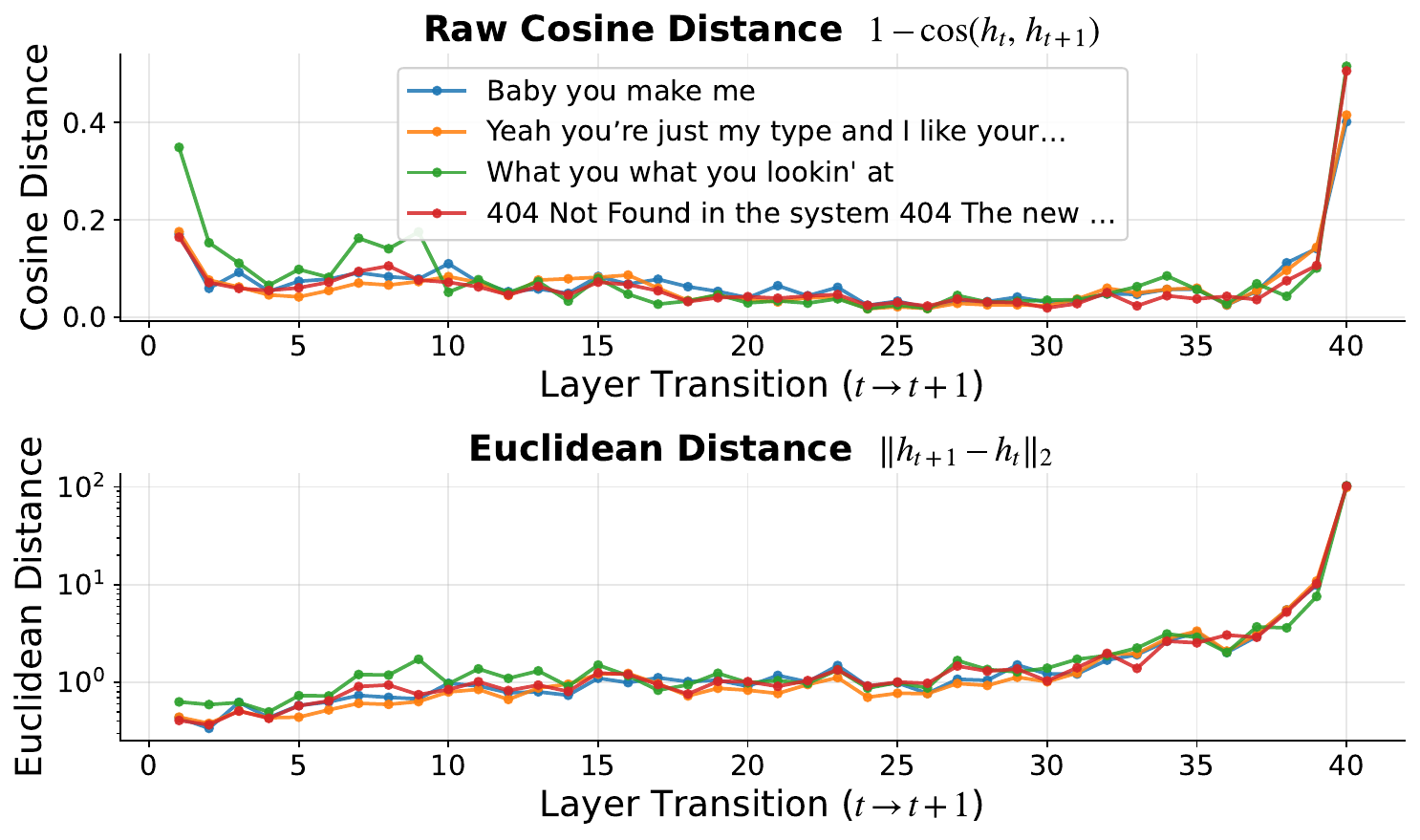}
	\caption{Qwen3.5-35B-A3B}
\end{subfigure}

\vspace{0.5em}

\begin{subfigure}[t]{0.49\textwidth}
	\centering
	\includegraphics[width=\textwidth]{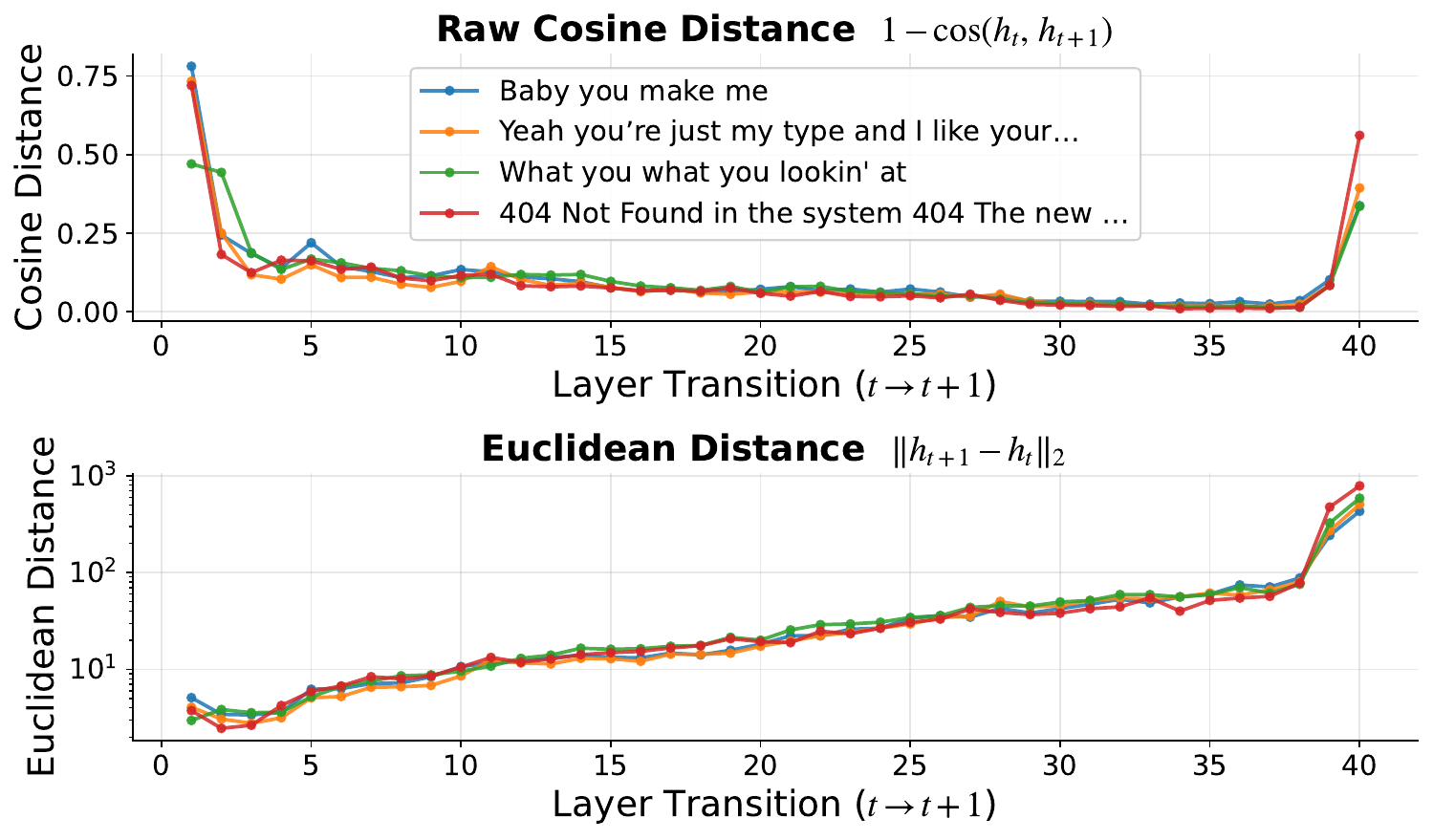}
	\caption{Phi-4}
\end{subfigure}
\hfill
\begin{subfigure}[t]{0.49\textwidth}
	\centering
	\includegraphics[width=\textwidth]{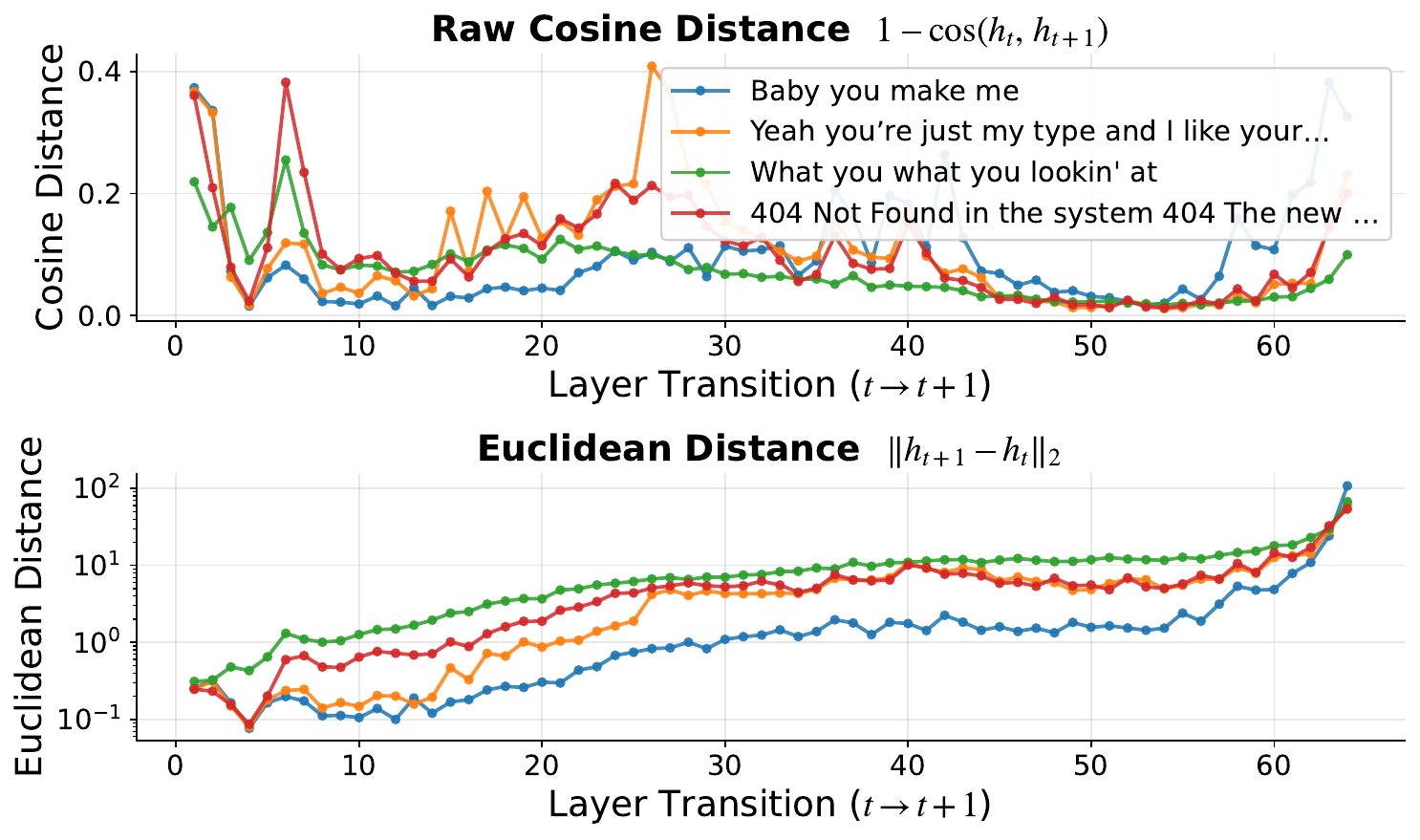}
	\caption{EXAONE-4.0-32B}
\end{subfigure}

\caption{Hidden-state distance profiles across model backbones.}
\label{fig:appendix_distance}
\end{figure}

To test whether the staged representation dynamics used to motivate LoopUS are specific to a single backbone or reflect a broader pattern, we extend the hidden-state analysis to additional pretrained LLMs. Figure~\ref{fig:appendix_pca} plots PCA trajectories of hidden states across layers for six backbones, while Figure~\ref{fig:appendix_distance} reports the corresponding layer-to-layer distance profiles. Together, these figures show that the qualitative structure highlighted in Figure~\ref{fig:hidden_analysis_original_LLM} is not unique to Qwen3-1.7B.

Across model families, we repeatedly observe the same three-phase organization: rapid representational motion in early layers, a relatively smooth middle-layer regime, and a sharper transition near the final layers. In the PCA plots, this appears as a gradual arc or plateau through the middle of the network followed by a more pronounced turn toward the output-facing layers. In the distance profiles, the same effect appears as reduced consecutive-layer change through the middle block bracketed by larger changes at the beginning and end of the network. This cross-model consistency supports our architectural prior that pretrained decoder-only transformers naturally admit an encoder--reasoning--decoder decomposition, with the middle layers forming the most suitable region for stable iterative reuse.

\section{Halting Strategies}
\label{appendix:halting_strategies}

\begin{figure}[h!]
\centering
\includegraphics[width=\textwidth]{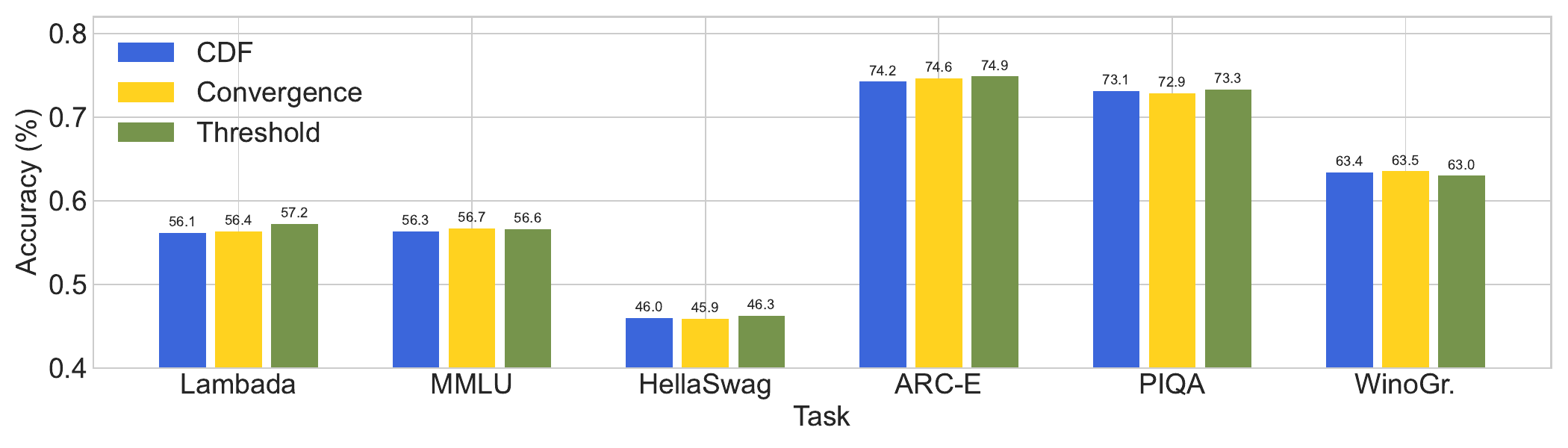}
\caption{Halting behavior under different stopping strategies. The learned threshold-based rule yields the sharpest halting distribution, indicating that LoopUS learns a meaningful stopping signal.}
\label{fig:halting_comparison}
\end{figure}

We compare three inference-time halting rules while keeping the LoopUS checkpoint fixed. Let $\bar{h}_{t+1}^{(b)} \in \mathbb{R}^{h}$ denote the hidden state of the newest token after the $b$-th latent refinement step at autoregressive decoding step $t+1$, and let
\begin{equation}
\tilde{q}_{t+1}^{(b)} = q_{\phi}\!\left(\bar{h}_{t+1}^{(b)}\right),
\qquad
q_{t+1}^{(b)} = \sigma\!\left(\tilde{q}_{t+1}^{(b)}\right).
\end{equation}
Given a maximum recursion budget $B$, each strategy returns an exit depth
\begin{equation}
b_{\mathrm{exit}}(t+1)
=
\min\!\left\{b \in \{1,\dots,B\}: \mathrm{Stop}_{t+1}^{(b)}=1\right\},
\end{equation}
with the convention that $b_{\mathrm{exit}}(t+1)=B$ if no earlier stopping condition is satisfied. In our implementation, the confidence head is evaluated at every refinement step by default ($\texttt{q\_eval\_interval}=1$), although the code also supports evaluating it every $r$ steps and always at the final step. During batched evaluation, halting is applied conservatively so that all examples in the active batch remain synchronized: threshold and CDF halting use the minimum stopping score across the batch, whereas convergence halting uses the maximum hidden-state change across the batch.

\paragraph{Threshold-based halting.}
This is the main stopping rule used by LoopUS and corresponds directly to the confidence-based stopping rule introduced in Section~\ref{sec:method_confidence_head}. We stop as soon as the confidence head exceeds a fixed threshold,
\begin{equation}
q_{t+1}^{(b)} \ge q_{\mathrm{th}},
\end{equation}
or equivalently,
\begin{equation}
b_{\mathrm{exit}}^{\mathrm{th}}(t+1)
=
\min\!\left\{b: q_{t+1}^{(b)} \ge q_{\mathrm{th}}\right\}.
\end{equation}
Because $\mathcal{L}_{Q}$ trains $q_{\phi}$ to approximate the post-update token accuracy, this rule interprets $q_{t+1}^{(b)}$ as a direct estimate that the current latent state is already sufficiently predictive and that further refinement is unlikely to be necessary. For a batch $\mathcal{B}$, the implementation stops only when
\begin{equation}
\min_{n \in \mathcal{B}} q_{n,t+1}^{(b)} \ge q_{\mathrm{th}},
\end{equation}
which ensures that every sequence in the batch is ready to halt before the shared forward pass terminates.

\paragraph{Convergence-based halting.}
The second strategy is a non-parametric baseline that monitors whether the latent trajectory has numerically converged. Let
\begin{equation}
\Delta_{t+1}^{(b)} = \left\|\bar{h}_{t+1}^{(b)} - \bar{h}_{t+1}^{(b-1)}\right\|_{2},
\qquad b \ge 1,
\end{equation}
where $\bar{h}_{t+1}^{(b-1)}$ and $\bar{h}_{t+1}^{(b)}$ are the last-token hidden states before and after the current refinement step. The model exits when the representation change becomes smaller than a preset tolerance $\epsilon$,
\begin{equation}
\Delta_{t+1}^{(b)} \le \epsilon,
\end{equation}
that is,
\begin{equation}
b_{\mathrm{exit}}^{\mathrm{conv}}(t+1)
=
\min\!\left\{b: \Delta_{t+1}^{(b)} \le \epsilon\right\}.
\end{equation}
This rule does not use the confidence head at all. Instead, it assumes that once the hidden state of the current token changes only marginally from one loop to the next, the decoder distribution is unlikely to improve enough to justify another refinement step. In batched evaluation, we stop only when
\begin{equation}
\max_{n \in \mathcal{B}} \Delta_{n,t+1}^{(b)} \le \epsilon,
\end{equation}
so that the entire batch has converged under the same geometric criterion.

\paragraph{CDF-based halting.}
For comparison, we also implement a cumulative-distribution-based halting rule inspired by the Q-exit criterion used in Ouro~\cite{zhu2025ouro}. In this strategy, the stepwise confidence is reinterpreted as a hazard rate,
\begin{equation}
\lambda_{t+1}^{(b)} = q_{t+1}^{(b)} \in (0,1).
\end{equation}
The probability of exiting \emph{exactly} at step $b$ is then defined by
\begin{equation}
\pi_{t+1}^{(b)}
=
\lambda_{t+1}^{(b)} \prod_{j=1}^{b-1} \left(1-\lambda_{t+1}^{(j)}\right),
\end{equation}
and the cumulative probability of having exited by step $b$ is
\begin{equation}
\mathrm{CDF}_{t+1}^{(b)}
=
\sum_{i=1}^{b} \pi_{t+1}^{(i)}
=
1 - \prod_{j=1}^{b} \left(1-\lambda_{t+1}^{(j)}\right).
\end{equation}
The corresponding stopping rule is
\begin{equation}
b_{\mathrm{exit}}^{\mathrm{cdf}}(t+1)
=
\min\!\left\{b: \mathrm{CDF}_{t+1}^{(b)} \ge q_{\mathrm{th}}\right\}.
\end{equation}
Intuitively, this rule accumulates evidence for stopping over multiple refinement steps: even if no single step produces a confidence score above $q_{\mathrm{th}}$, several moderately large values of $\lambda_{t+1}^{(b)}$ can combine into a large cumulative exit probability. For numerical stability, the implementation accumulates the survival probability in log space,
\begin{equation}
\log S_{t+1}^{(b)} = \sum_{j=1}^{b} \log\!\left(1-\lambda_{t+1}^{(j)}\right),
\qquad
\mathrm{CDF}_{t+1}^{(b)} = 1 - \exp\!\left(\log S_{t+1}^{(b)}\right).
\end{equation}
Since LoopUS trains its confidence head with a direct binary target on post-update token accuracy rather than with Ouro's full exit-distribution objective, we use this CDF rule as a comparative inference heuristic rather than as the primary training-consistent stopping criterion. In batched evaluation, stopping occurs only when
\begin{equation}
\min_{n \in \mathcal{B}} \mathrm{CDF}_{n,t+1}^{(b)} \ge q_{\mathrm{th}}.
\end{equation}

\paragraph{Discussion.}
These three strategies expose different inductive biases. Threshold halting is the rule most tightly matched to our training objective because $q_{\phi}$ is explicitly supervised to predict whether the current refinement is already good enough. Convergence halting ignores the confidence head and instead uses only latent geometry, making it a useful diagnostic for whether the loop behaves like a contractive refinement process. The CDF-based rule is more aggressive in aggregating multiple moderate confidence values across steps and therefore serves as a natural comparison point to prior looped-language-model early-exit mechanisms~\cite{zhu2025ouro}. Figure~\ref{fig:halting_comparison} compares these strategies under matched checkpoints while varying $q_{\mathrm{th}}$ or $\epsilon$.

\section{Limitations and Future Work}
\label{sec:limitations_and_future_work}

\paragraph{Extension beyond text-only language modeling.}
All experiments in this paper focus on text-only, decoder-only language models. Although LoopUS is formulated as a latent-space refinement method and is therefore not tied to a particular input modality in principle, we have not evaluated whether the same encoder--reasoning--decoder decomposition remains stable for multimodal models. Recent progress on recurrent and looped architectures suggests that iterative latent computation can also be useful beyond text-only settings, but it remains unclear whether a pretrained multimodal transformer exhibits the same middle-layer geometry that makes LoopUS effective in our experiments~\cite{xu2026looping,zhu2025surveylatentreasoning}. Vision, audio, and cross-modal tokens may induce different hidden-state structure, different layer specialization, and different failure modes under repeated latent refinement. Extending LoopUS to vision-language and other multimodal transformers is therefore an important next step, especially for testing whether looped latent computation can improve grounding, compositional perception, and cross-modal reasoning rather than only text-domain prediction.

\paragraph{Heterogeneous and hybrid model architectures.}
Our experiments instantiate LoopUS on relatively standard transformer backbones. Modern language models increasingly combine heterogeneous layers and operators, such as gated delta networks~\cite{qwen3.6-27b}, sparse attention~\cite{deepseekai2026deepseekv4}, mixture-of-experts routing~\cite{choi2026kexaonetechnicalreport,park2026solaropentechnicalreport}, state-space modules~\cite{nvidia2025nvidianemotron3efficient}, and other hybrid sequence mechanisms~\cite{kimiteam2026attentionresiduals}. In such models, the middle-layer region may no longer behave as a uniform reusable block, and the optimal recursion policy may depend on the operator type, layer role, or token state. Future work should study how to select, compose, or schedule looped modules in architectures whose layers have substantially different computational semantics. A particularly useful direction is to learn architecture-aware recursion policies that decide not only \emph{when} to stop, but also \emph{which} operator or layer group should be reused at each refinement step.

\paragraph{Scaling to larger and more diverse training regimes.}
The present results are obtained under a moderate post-training budget rather than a full large-scale training pipeline. This setting is useful for isolating the effect of looped up-scaling, but it leaves open how LoopUS behaves under substantially larger corpora, longer contexts, curriculum schedules, or more diverse data mixtures. Scaling the training recipe is especially relevant because the confidence head and monotonicity objective may benefit from richer distributions of refinement trajectories than those observed in limited-token adaptation. Larger-scale studies could also clarify whether the observed gains remain concentrated on reasoning-oriented tasks or broaden to knowledge-heavy and long-context settings as the loop is exposed to more varied supervision.

\paragraph{Dedicated math and long-context reasoning coverage.}
Although our benchmark suite includes several reasoning-oriented evaluations, it does not include dedicated mathematical reasoning tasks or training on math-focused corpora. This is an important limitation of the current study. Because of compute constraints, our released training setup remained at a context length of 1024 and did not incorporate additional math-oriented datasets during adaptation. As a result, the paper does not yet test whether LoopUS yields similar gains on problems that require longer derivations, sustained multi-step reasoning, or explicit mathematical supervision. A useful next step is therefore to scale the training budget and context window while adding dedicated math and reasoning datasets, so that the benefits and failure modes of latent looping can be assessed under more computation-intensive reasoning workloads.

\paragraph{Integration with instruction tuning and preference optimization.}
We evaluate LoopUS as a post-training adaptation method for base models. In practical LLM development, however, model quality is shaped not only by continued pretraining but also by instruction tuning, long-context adaptation, and preference optimization methods such as reinforcement learning from feedback or GRPO-style objectives~\cite{tie2025surveyposttraininglargelanguage}. Integrating LoopUS with SFT and preference-based post-training could test whether looped latent computation improves not only benchmark accuracy but also instruction following, controllability, calibration, and user-facing reliability~\cite{zhu2025scaling}. This direction is also important because adaptive test-time computation may interact with alignment objectives: a model should learn when extra refinement is helpful for faithfully following an instruction and when additional computation risks overthinking or drifting from the user's intent.

\paragraph{LoopUS applied to diffusion LLMs.}
Although our current implementation focuses on standard autoregressive transformers, the core mechanism of LoopUS---iterative latent refinement controlled by dynamic gating and early halting---naturally aligns with the continuous denoising trajectories of diffusion language models~\citep{nie2025large,zhou2026dllmsimplediffusionlanguage}. Diffusion LLMs fundamentally rely on iterative processes to refine continuous latent representations before decoding. However, their denoising steps are typically fixed or bound to computationally uniform schedules. Integrating the LoopUS framework into diffusion LLMs could introduce data-dependent adaptive computation to the reverse diffusion process, allowing the model to halt denoising dynamically when the latent state is sufficiently clean, guided by our confidence head. Furthermore, the selective gating mechanism and monotonicity loss could serve as structural regularizers for the score network, ensuring stable, monotonic improvement across timesteps and potentially accelerating generation without requiring explicit trajectory distillation. Exploring this intersection remains a promising direction for making continuous-space language generation more practical.

\paragraph{LoopUS as a pretraining-native architecture.}
Finally, LoopUS may also be useful earlier in the model lifecycle. This paper treats looped depth up-scaling as a minimally invasive retrofit applied to pretrained backbones, but the same architectural prior could be incorporated during pretraining itself. Pretraining with looped latent refinement from the beginning may allow the model to allocate representational capacity differently, learn more calibrated stopping behavior, and make better use of adaptive test-time computation. A pretraining-native LoopUS model could also expose the recurrence to a much broader distribution of intermediate states, potentially reducing the gap between training-time refinement and inference-time recursion.


\end{document}